\documentclass[10pt,twocolumn,letterpaper]{article}

\usepackage{cvpr}      

\usepackage[dvipsnames]{xcolor}

\usepackage[accsupp]{axessibility}
\definecolor{cvprblue}{rgb}{0.21,0.49,0.74}
\usepackage[pagebackref,breaklinks,colorlinks,citecolor=cvprblue]{hyperref}

\def\paperID{14165} 
\def\confName{CVPR}
\def\confYear{2024}

\usepackage{textcomp}  
\usepackage{scalerel}  

\title{Spin-UP: Spin Light for Natural Light Uncalibrated Photometric Stereo}

\author{
Zongrui~Li$^{1,2,}$\thanks{Co-first author. ~~\ddag Corresponding author.}~
Zhan~Lu$^{2,4,*,}$\thanks{Work completed while interning at the State Key Lab of Brain-Machine Intelligence, Zhejiang University.}~
Haojie~Yan$^{3,4}$
Boxin~Shi$^{5,6}$
Gang~Pan$^{3,4}$
Qian Zheng$^{3,4,\ddag}$
Xudong Jiang$^{1,2}$
\smallskip 
\\
$^1${\small Rapid-Rich Object Search (ROSE) Lab, Interdisciplinary Graduate Programme, Nanyang Technological University}\\
$^2${\small School of Electrical and Electronic Engineering, Nanyang Technological University}\\
$^3${\small College of Computer Science and Technology, Zhejiang University}\\
$^4${\small The State Key Lab of Brain-Machine Intelligence, Zhejiang University}\\
$^5${\small National Key Laboratory for Multimedia Information Processing, School of Computer Science, Peking University}\\
$^6${\small National Engineering Research Center of Visual Technology, School of Computer Science, Peking University}\\
{\small\{zongrui001,~zhan007,~EXDJiang\}@ntu.edu.sg},
{\small \{hjyan,~gpan,~qianzheng\}@zju.edu.cn},
{\small shiboxin@pku.edu.cn}
}

\def\eg{\emph{e.g.}} 
\def\ie{\emph{i.e}} 
\newcommand{\Tref}[1]{Table~\ref{#1}}
\newcommand{\Eref}[1]{Eq.~(\ref{#1})}
\newcommand{\Fref}[1]{Fig.~\ref{#1}}
\newcommand{\Sref}[1]{Sec.~\ref{#1}}
\newcommand{\Aref}[1]{Algorithm~\ref{#1}}

\usepackage{multirow}
\usepackage{nicematrix}
\usepackage{amsmath}
\usepackage{amssymb}
\usepackage[utf8]{inputenc} 
\usepackage[T1]{fontenc}    
\usepackage{url}            
\usepackage{booktabs}       
\usepackage{amsfonts}       
\usepackage{nicefrac}       
\usepackage{graphicx}
\usepackage{amsmath}
\usepackage{amssymb}
\usepackage{makecell}
\usepackage{algorithmicx}
\usepackage{algorithm}
\usepackage[noend]{algpseudocode}
\usepackage{wrapfig,lipsum,booktabs}
\usepackage{threeparttable}

\begin{document}
\maketitle

\begin{abstract}
Natural Light Uncalibrated Photometric Stereo (NaUPS) relieves the strict environment and light assumptions in classical Uncalibrated Photometric Stereo (UPS) methods. However, due to the intrinsic ill-posedness and high-dimensional ambiguities, addressing NaUPS is still an open question. Existing works impose strong assumptions on the environment lights and objects' material, restricting the effectiveness in more general scenarios. Alternatively, some methods leverage supervised learning with intricate models while lacking interpretability, resulting in a biased estimation. In this work, we propose \textbf{Spin} Light \textbf{U}ncalibrated \textbf{P}hotometric Stereo (\textbf{Spin-UP}), an unsupervised method to tackle NaUPS in various environment lights and objects. The proposed method uses a novel setup that captures the object's images on a rotatable platform, which mitigates NaUPS's ill-posedness by reducing unknowns and provides reliable priors to alleviate NaUPS's ambiguities. Leveraging neural inverse rendering and the proposed training strategies, Spin-UP recovers surface normals, environment light, and isotropic reflectance under complex natural light with low computational cost. Experiments have shown that Spin-UP outperforms other supervised / unsupervised NaUPS methods and achieves state-of-the-art performance on synthetic and real-world datasets. Codes and data are available at 
\textcolor{magenta}{https://github.com/LMozart/CVPR2024-SpinUP}.
\end{abstract}
\vspace{-10pt}
\section{Introduction}
Natural light uncalibrated photometric stereo (NaUPS)~\cite{guo2021patch} is proposed to relieve the dark room and directional light assumption in classical uncalibrated photometric stereo, aiming to reconstruct the surface normal given images of an object captured at arbitrary environment light. The implications of NaUPS are far-reaching: it makes photometric stereo universal. However, solving NaUPS is still an open question because of the intrinsic ill-posedness introduced by the varying light of each image and the high dimensional ambiguities between the light and objects~\cite{guo2021patch}. 

Previous optimization-based methods use the simple light model to represent the varying environment lights and Lambertian reflectance to represent the material~\cite{guo2021patch, haefner2019variational, queau2015tv}. These models help mitigate the ill-posedness and ambiguities to some extent but become ineffective in handling objects with general reflectance (\eg, non-Lambertian reflectance) under complex environment light, leading to unsatisfactory reconstruction outcomes. Besides, since they solely model the varying lights in each image, the unknowns introduced by the light model may increase with the resolution and numbers of the images, restricting their method to low-resolution and insufficient images.

Considering the difficulties of explicitly mitigating the ill-posedness and ambiguities, recent advances~\cite{ikehata2022universal, ikehata2023scalable} turn to data-driven methods: they train a deep learning model on large-scale datasets and implicitly exploit deep light features from images to improve performance. However, those methods lack interpretability, making them hard to constrain during training. Consequently, the model may be affected by the data bias and prone to specific types of light sources~\cite{ikehata2022universal} or reflection variations~\cite{ikehata2023scalable} among images.

Despite a persistent exploration in this research field, a method capable of handling general objects under natural light while free from data bias is still missing. In this paper, we provide a new perspective on solving NaUPS. Specifically, we propose a novel setup that acquires images on a rotatable platform under a static environment light. In such a case, the object is illuminated by rotated environment lights. Representing such rotated lights requires fewer parameters than solely modeling varying lights in each image since we model the lights as a uniform environment light varied by low degree-of-freedom (DOF~\cite{gardner2022rotation}) rotation. This frees us from significantly increasing unknowns when implementing advanced parametric light models (\eg, spherical Gaussian) and reflectance models to handle general scenarios.  Additionally, based on such a setup, we further derive a reliable light initialization method by analyzing the pixel value at object's occluding boundary. Such light initialization will help the model converge at the beginning and alleviate the ambiguity between light and objects during training. 

With the help of the proposed setup and light initialization method, we develop \textbf{Spin} Light \textbf{U}ncalibrated \textbf{P}hotometric Stereo (\textbf{Spin-UP}), addressing NaUPS by optimizing inverse rendering framework in an unsupervised manner. To our best knowledge, this is the first unsupervised method that can handle general objects under natural light. Unlike previous methods, Spin-UP can jointly reconstruct environment light, isotropic reflectance, and complicated shapes through iterative optimization. To reduce the computational cost and improve convergence, we propose two strategies: interval sampling and shrinking range computing, which help optimize Spin-UP in low GPU usage (5 GB) and fast running time (25 mins). Experiments on synthetic and real-world datasets demonstrate our superior performance over previous methods on general scenarios. 
Overall, our contributions are summarized as follows:

\begin{enumerate}[itemsep=0.1pt,topsep=0.1pt,parsep=0.1pt]
    \item We design a novel setup for NaUPS, which reduces unknowns of light representation and facilitates solving NaUPS in an unsupervised manner. 
    \item  We introduce a light prior, which leverages an object’s occluding boundaries to initialize a reliable environment light. Based on the setup and light prior, we propose the unsupervised NaUPS method named Spin-UP.
    \item We present two training strategies for fast training and convergence of Spin-UP. 
\end{enumerate}
\section{Related Work}
In this section, we briefly review recent supervised and unsupervised NaUPS methods. We also summarize other techniques that exploit priors from occluding boundaries. Additionally, we discuss recent advances in 3D vision to distinguish Spin-UP from other neural inverse rendering approaches. Given that the focus of this paper is on NaUPS, a group of works reconstructing 3D surfaces from a single image by deep learning under natural light~\cite{li2018learning} or shading~\cite{durou2008numerical, zhang1999shape, johnson2011shape} are not included.

\noindent \textbf{Natural Light Uncalibrated Photometric Stereo}. Unsupervised NaUPS methods jointly recover the light, reflectance properties, and surface normal. These methods explicitly model the environment light by low-order spherical harmonics (SH)~\cite{basri2007photometric, queau2015tv}, spatially varying spherical harmonics (SV-SH)~\cite{haefner2019variational, maier2017intrinsic3d} or equivalent directional light (Eqv-Dir)~\cite{guo2021patch} to mitigate the ill-posedness, and use integrability constraint~\cite{basri2007photometric}, shape initialization~\cite{haefner2019variational}, non-physical lighting regularization~\cite{queau2015tv}, or graph-based ambiguity relaxation~\cite{guo2021patch} to alleviate the ambiguity. In contrast, supervised NaUPS methods~\cite{ikehata2022universal, ikehata2023scalable} apply deep learning models like transformers to reconstruct normal maps without explicitly estimating the environment light. The models are trained on a dataset containing images of diverse objects captured under various lighting conditions, including directional, point, and environment light.
Compared to previous work, the proposed Spin-UP distinguishes itself in three key aspects: 1) it features a novel setup explicitly designed to model correlations among observed images to mitigate the ill-posedness of NaUPS, 2) leveraging this unique setup, a novel light initialization method is introduced to mitigate ambiguities, and 3) an advanced light and material model is implemented to address a broader range of scenarios. 

\noindent \textbf{Priors from the Boundaries}. The occluding boundaries of an object are considered to reveal adequate information about the object's shape and the scene's light. Given the fact that the projection of the boundaries' normal to xy-plane is perpendicular to the boundaries in orthographic projection, methods are developed to constraint the surface normal estimation during iterative optimization~\cite{li2023dani} or recover a rough shape to initialize the geometry in multi-view~\cite{hernandez2008multiview} or photometric stereo~\cite{haefner2019variational}. Other methods associate the boundaries normal with the reflectance to estimate a rough position of the directional lights~\cite{vogiatzis2006reconstruction}. However, none of them derive the environment light from the boundary reflectance. Given the setup in Spin-UP, we can roughly estimate the environment by analyzing occluding boundaries and the corresponding pixel points. This approach provides a reliable initialization for lighting that alleviates ambiguity in NaUPS.

\noindent \textbf{Inverse rendering in 3D Vision}. Neural Radiance Fields (NeRF)~\cite{mildenhall2020nerf} implicitly store the scene's shape and reflectance through MLPs optimized by inverse volume rendering. While NeRF can only recover coarse 3D shapes, several subsequent works~\cite{zhang2021physg,wang2021neus,verbin2021ref,zhang2021nerfactor,Boss_2021_ICCV,Srinivasan_2021_CVPR} have been proposed to combine the surface rendering and volume rendering techniques, recovering fine shapes under varying viewpoints but static environment light. In contrast, viewpoints in Spin-UP are relatively static to the objects.  While most neural field methods aim to recover the whole 3D geometries, Spin-UP only recovers the object's surface.
\vspace{-5pt}

\section{Proposed Method}
In~\Sref{sec:spinlight}, we explain the Spin-UP's setup and how it reduces the unknowns. In~\Sref{sec: Spin-UP}, we introduce the light prior that alleviates ambiguities in NaUPS, including details of the light initialization method based on that prior. In~\Sref{sec: opt spin-up}, we describe the implementation details of the proposed Spin-UP framework and losses. In~\Sref{sec:ts}, we demonstrate two proposed training strategies.

\begin{figure}[t]
    \centering
    \includegraphics[width=0.47\textwidth]{"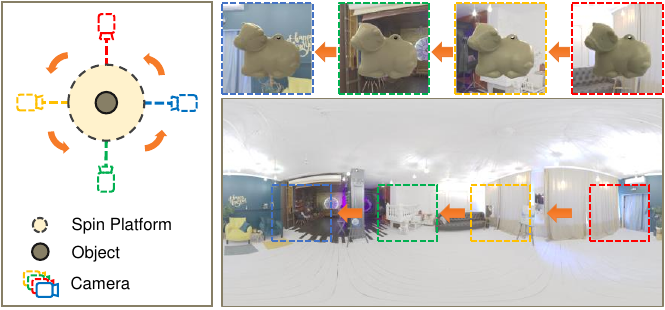"} 
    \caption{The proposed image capturing setup. Left: an illustration of image-capturing equipment consisting of a rotatable platform, a camera, and the target object. We spin the platform in $360^\circ$ and capture images of the object. The object and camera rotate together with the platform. Right-top: Four observed images at different positions. Right-bottom: Ground truth environment light. Dashed color boxes indicate the corresponding camera views. } 
    \label{fig:setup}
    \vspace{-10pt}
\end{figure}

\subsection{Spin Light Setup}
\label{sec:spinlight}

As shown in \Fref{fig:setup}, we capture a sequence of images $\boldsymbol{I} \triangleq \{I_j | j\in [1, ..., N_I]\}$ for an object\footnote{We assume the object's boundary is geometrically smooth (occluding boundary).} by rotating it together with a linear perspective camera in $360^\circ$ on a rotatable platform.
Since the relative positions and orientations between the camera and the object are fixed during rotation, each observed image is aligned with the rotated environment light $\boldsymbol{L}(\boldsymbol{R}_j \cdot \boldsymbol{\omega})$, where $\boldsymbol{\omega} \in \mathbb{R}^3$ indicates the incident light's direction, $\boldsymbol{R}_j=\boldsymbol{R}(\theta_j)$ is the 1-DoF~\cite{gardner2022rotation} rotation matrix representing rotation about the vertical axis, $\theta_j$ is the rotation angle, $\theta_1=0$. As we conduct $360^{\circ}$ rotation and assume a constant velocity (despite not strictly required in practice), $\boldsymbol{R}_j$ can be initialized given $\theta_j = 2\pi (j-1) / {N_I}$. Given a sequence of images $\boldsymbol{I}$ and the initialized $\boldsymbol{R}$, Spin-UP iteratively optimizes the normal map $\boldsymbol{N}$, the environment light $\boldsymbol{L}$, isotropic BRDF map $\boldsymbol{M}$, and rotation angle $\boldsymbol{R}$ by solving
{\setlength\abovedisplayskip{0.1cm}
\setlength\belowdisplayskip{0.1cm}
\begin{align}
\label{eq:inv_rend}
    \underset{\boldsymbol{L}, \boldsymbol{M}, \boldsymbol{N}, \boldsymbol{R}}{\arg \min } \sum_{i=1}^{N_P} \sum_{j=1}^{N_I} \mathcal{E}\left(\boldsymbol{m}_{ij}, \hat{\boldsymbol{m}}_{ij}\right),
\end{align}}where $N_P$ is the number of sampled points on the surface, $\boldsymbol{m}_{ij}$ and $\hat{\boldsymbol{m}}_{ij}$ indicate the ground truth and estimation of point $i$'s color in image $I_j$, respectively. 
$\mathcal{E}(\cdot,\cdot)$ is loss function between $\boldsymbol{m}_{ij}$ and $\hat{\boldsymbol{m}}_{ij}$ (\ie, mean absolute error).
We adopt the rendering equation to calculate the color $\boldsymbol{\hat{m}}$\footnote{The subscripts are omitted for simplicity.}
{\setlength\abovedisplayskip{0.1cm}
\setlength\belowdisplayskip{0.1cm}
\begin{align}
\begin{aligned}
\label{eq:render_eq}
    \boldsymbol{\hat{m}} &=\int_\Omega s\boldsymbol{L}\left(\boldsymbol{\omega}\right) \rho\left(\boldsymbol{\omega} \cdot \boldsymbol{n}\right) \mathrm{d} \boldsymbol{\omega}, \\
     & =\int_\Omega s\boldsymbol{L}\left(\boldsymbol{\omega}\right) (\rho^s + \rho^d)\left(\boldsymbol{\omega} \cdot\boldsymbol{n}\right) \mathrm{d} \boldsymbol{\omega}.
\end{aligned}
\end{align}}where $\Omega$ represents the upper hemisphere centered at the normal vector $\boldsymbol{n}$. $s$ is the cast shadow.
$\rho^s$ and $\rho^d$ indicate the specular and diffuse reflectance, respectively. The ambiguities between the light $\boldsymbol{L}$ and reflectance of the object $\boldsymbol{M}$ are often disregarded~\cite{li2023dani, li2022self}.

\begin{table}[t]
\setlength{\abovecaptionskip}{0.1cm}
\setlength{\belowcaptionskip}{0.1cm}
\setlength{\tabcolsep}{15pt}
\caption{A comparison of type, unknowns' number (number), and representation capacity (rep. capacity) of light model among Spin-UP and representative unsupervised NaUPS methods. The unknown number is calculated on $50$ images ($512\times512$). `freq.' represents frequency.}
    \label{tab:method}
    \centering
    \vspace{0pt}
    \resizebox{1\linewidth}{!}{
    \begin{tabular}{cccccc}
    \hline
      & QL15~\cite{queau2015tv} & HY19~\cite{haefner2019variational} & GM21~\cite{guo2021patch} & Spin-UP\\
    \midrule
    type  & SV-SH & Global SH  & Eqv-Dir & SG\\
    number   & $1.5$K & $450$  & $1.4$M & $434$\\
    \makecell{rep. capcity}  & low-freq. & low-freq. & low-freq.  & high-freq.\\
    \hline
    \end{tabular}
    }
\end{table}

\begin{figure}[t]
    \centering
    \includegraphics[width=\linewidth]{"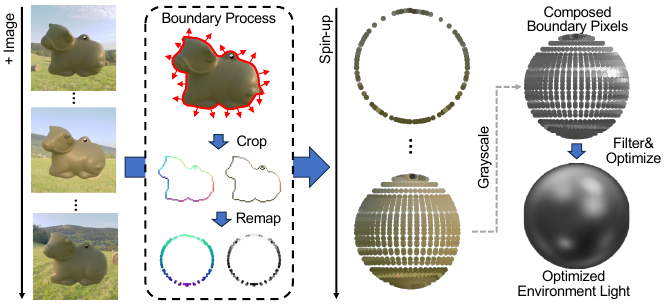"} 
    \caption{The proposed light initialization method in Spin-UP. 
    We crop the boundary pixels $m_b$ and normal $\boldsymbol{n}_b$ from input images. 
    Then, we remap them on the sphere and rotate them with their corresponding rotations $\boldsymbol{R}$.
    Based on a light probe composed of gray-scale boundary pixels, we optimize the SG light model to obtain the environment light.}
    \label{fig:light_probe}
    \vspace{-5pt}
\end{figure}  



\noindent \textbf{Unknowns reduction.} 
The proposed spin light setup reduces the unknowns of the light representation $\boldsymbol{L}$ by exploiting correlations between different images. 
Unlike previous NaUPS methods that separately model the light for each image, we consider a uniform environment light $\boldsymbol{L}$ for all images represented by the parametric model like spherical Gaussian, and 1-DoF rotation angle $\theta$ for each image. 
As such, the unknowns consist of the environment light model's parameters and the number of rotational angles that are quantitatively equal to $N_I$. The total unknown amount is reduced compared to other methods (\Tref{tab:method}), which helps mitigate the ill-posedness and facilitates solving NaUPS with advanced light and reflectance models in an unsupervised manner.

\subsection{Light Prior from Boundaries}
\label{sec: Spin-UP}

Based on the spin light setup, we can exploit priors from the object boundary for light initialization to alleviate the ambiguity.
The idea is motivated by the observation that the pixel value $m_b$ at an object’s boundary provides insights into the environment light (see \Fref{fig:light_probe}). For an object with occluding boundaries, the normal of those boundaries $\boldsymbol{n}_b$ can be pre-computed~\cite{li2022self, li2023dani}.  By bonding $m_b$, $\boldsymbol{n}_b$, and $\boldsymbol{R}$, we can roughly derive a light map indicating the light sources' positions and intensities, where $m_b$ directly represent the light intensity $\boldsymbol{L}(\boldsymbol{\omega}_b)$ at $\boldsymbol{\omega}_b = \boldsymbol{R}\cdot\boldsymbol{n}_b$. However, the derived light map for objects with different materials may contain mismatched light source positions and chromatic bias, leading to inaccurate light initialization. 

\begin{figure}[t]
    \centering
    \includegraphics[width=0.4\textwidth]{"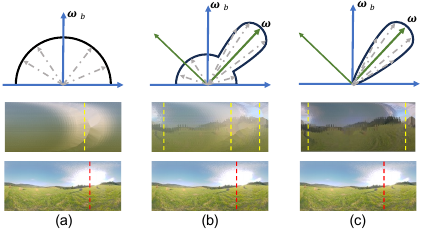"} 
    \caption{An illustration of mismatched light source positions. Rows from top to bottom: an illustration of reflection on materials with different roughness; the initial environment light before applying any filters given objects' boundary pixels; the ground truth environment light. Objects are (a) the diffuse-dominant sphere, (b) the diffuse and specular mixture sphere, and (c) the specular dominant sphere. The yellow and red lines in rows 2 and 3 indicate a rough position of the light sources. } 
    \label{fig:light_comp}
    \vspace{-15pt}
\end{figure}

The mismatched light source positions are caused by the specular component $m^s_b$ in $m_b$. When $m_b^s$ dominates, approximating $\boldsymbol{L}(\boldsymbol{\omega}_b)$ by $m_b$ becomes a biased estimation as $m_b^s$ is a reflection of lights in different directions than $\omega_b$~(\Fref{fig:light_comp}). By contrast, the approximation is more reasonable when the diffuse component ($m_b^d$) dominates since $m^d_b=\int_{\Omega} \boldsymbol{L}\left(\boldsymbol{\omega}\right) \rho^d\left(\boldsymbol{\omega} \cdot \boldsymbol{\omega}_b\right) \mathrm{d} \boldsymbol{\omega}$ indicates that $\boldsymbol{L}(\boldsymbol{\omega}_b)$ contributes most to the actual pixel value, making it less biased to use $m^d_b$ to represent $\boldsymbol{L}(\boldsymbol{\omega}_b)$. Therefore, to facilitate a less biased estimation of environment light for initialization, a diffuse filter $\mathcal{F}^d(.)$ is necessary on $m^d_b$, alleviating the mismatched light source positions issue. 
Similarly, a chromatic filter $\mathcal{F}^c(.)$ is also required to reduce the chromatic bias caused by the spatially varying material at boundaries. 
The filtered pixel value $\hat{m}_b^d=\mathcal{F}^c (\mathcal{F}^d\left(m_b\right))$ are the basis for our light initialization method.

\noindent \textbf{Light initialization method.} The procedure of light initialization method is summarized in~\Aref{alg:init}. This method aims to derive an initial environment light model with parameter $\Theta$. Specifically, we use $N_L=64$ spherical Gaussian (SG) bases~\cite{wang2009all} as the light model, where $\boldsymbol{L}(\boldsymbol{\omega}|\boldsymbol{\xi}_t, \lambda_t, \boldsymbol{\mu}_t) = \sum^{N_L}_{t=1} G(\boldsymbol{\omega}; \boldsymbol{\xi}_t, \lambda_t, \boldsymbol{\mu}_t)$. $\boldsymbol{\xi}_t$, $\lambda_t$, and $\boldsymbol{\mu}_t$ stands for Gaussian lobes' direction, sharpness, and amplitude, respectively\footnote{In practice, we find that initializing Gaussians' parameters by Fibonacci lattice~\cite{zhang2021physg} and freezing $\lambda_t$ gives the best results.}. Inspired by~\cite{shi2013bi} indicating that diffuse reflectance can be approximated by the low-frequency reflectance, we design $\mathcal{F}^d(.)$ as a combination of a threshold filter $\mathcal{F}^d_{TH}(.)$ that removes high-intensity reflectance (top $20\%$ bright) based on boundary points' intensity profile~\cite{shi2013bi} and a low-pass filter (\ie, 3-order spherical harmonics filter\footnote{We implement a Gaussian filter and rescale the pixel value's range to $\mathcal{F}^d_{TH}(m_b)$'s range to suppress ringing effect and negative energy in SH.}) noted as $\mathcal{F}^d_{SH}$~\cite{shi2013bi, seeley1966spherical}. While $\mathcal{F}^d_{TH}(.)$ reduces the bias by removing the brightest parts that are usually aligned with specular reflectance in observed images. $\mathcal{F}^d_{SH}(.)$ helps extract the low-frequency reflectance. For $\mathcal{F}^c(.)$, we design it as a converter transferring pixel values into gray-scale\footnote{We use gray-scale value to initialize the RGB value of light model.}, mitigating biases from spatially varying material.
\vspace{-5pt}
\begin{algorithm}
\small
\caption{Light Initialization Method}\label{alg:init}
\begin{algorithmic}[t]
\State \textbf{Input}: pre-computed boundary normal $\boldsymbol{n}_b$, initial relative rotation matrices $\{\boldsymbol{R}_1\, ..., \boldsymbol{R}_{N_I}\}$, boundary pixels $\{{m_b}_1, ..., {m_b}_{N_I}\}$, diffuse filters $\mathcal{F}^d_{TH}(.)$ and $\mathcal{F}^d_{SH}(.)$, chromatic filter $\mathcal{F}^c(.)$, fitting epochs $N$, environment light with parameters $\boldsymbol{L}(.|\Theta)$, learning rate $\eta$.
\State \textbf{Output}: an initial environment light model $\boldsymbol{L}(.|\Theta^\prime)$.
\For{$j=1 ,..., N_I$}
    \State ${\boldsymbol{\omega}_b}_j = \boldsymbol{R}_j \cdot \boldsymbol{n}_b$
\EndFor

\State $\hat{m}^d_b$ = $\mathcal{F}^c(\mathcal{F}^d_{SH}(\mathcal{F}^d_{TH}(m_b)))$
\For{$e=1 ,..., N$}
\State $J = \sum (\boldsymbol{L}(\boldsymbol{\omega}_b | \Theta) - \hat{m}^d_b)^2$
\State $\Theta \leftarrow \Theta - \eta \partial J / \partial \Theta$
\EndFor
$\Theta^\prime = \Theta$
\end{algorithmic}
\end{algorithm}
\vspace{-10pt}

\subsection{Framework of Spin-UP}
\label{sec: opt spin-up}

With reliable initial SG lights and rotation matrices $\boldsymbol{R}$, we develop Spin-UP based on neural inverse rendering~\cite{zhang2021physg,Boss_2021_ICCV, li2023dani, li2022self} given the rendering equation in \Eref{eq:render_eq}. 

\noindent\textbf{Shape model}. We use the neural depth field to represent the 3D surface. A multi-layer perceptron (MLP) predicts the depth value given the image coordinates. To compute normal given the depth map, we extend the normal fitting method described in~\cite{li2023dani} to the perspective projection\footnote{Please refer to the supplementary material for more details about the network structures and modified normal fitting method.}.

\noindent\textbf{Material model}. We represent the spatially varying and isotropic reflectance as a modified Disney Model~\cite{zhang2021physg}. The diffuse albedo $\boldsymbol{\rho}^d$ is predicted by another MLP with a similar structure given the image coordinate. The spatially varying specular reflectance is calculated as a weighted sum of $N_S=12$ SG bases, so
$\boldsymbol{\rho}^s=\sum_{n=1}^{N_S} c^n\mathcal{D}\left(\boldsymbol{v},\boldsymbol{\omega}\right)\mathcal{F}\left(\boldsymbol{h}, \boldsymbol{\omega}\right) \mathcal{G}(\boldsymbol{n}, \boldsymbol{\omega}, \boldsymbol{v}, \lambda_n)$,
where $\mathcal{D}$, $\mathcal{F}$, and $\mathcal{G}$ accounts for micro-facet's normal distribution, Fresnel effects, and self-occlusion, respectively. $\boldsymbol{v}$ is the view direction. $\boldsymbol{h}$ is the half-vector, calculated by $\boldsymbol{h}=(\boldsymbol{v}+\boldsymbol{\omega}) / \left\|\boldsymbol{v}+\boldsymbol{\omega}\right\|$, $\lambda_n$ is the roughness terms initialized as $(0.1 + 0.9(n-1)) / (N_S-1)$ and set as learnable parameters, $c^n$ is the weights predicted by the MLP.

\noindent\textbf{Shadow model}. We apply a shadow mask similar in~\cite{li2022self} to handle the cast shadow.

\noindent \textbf{Loss functions}. Similar to other inverse rendering-based methods~\cite{li2022self,li2023dani}, we use the inverse rendering loss (\ie, \Eref{eq:inv_rend}) to train the framework. The three-stage schema~\cite{li2023dani} is applied, as well as other smoothness terms (total variance regularization~\cite{li2022self, li2023dani}) calculate as
$\operatorname{TV}(.) = \frac{1}{N_P} \sum_{i=1}^{N_P}\left|\frac{\partial (.)}{\partial x}+\frac{\partial (.)}{\partial y}\right|$, where $(x, y)$ is the image coordinates. We implement $\operatorname{TV}(.)$
on the normal map ($\boldsymbol{N}$), diffuse albedo map ($\boldsymbol{A}$), and the Gaussian bases’ weights ($c$) for the material, and gradually drop it following the three-stage training schema~\cite{li2023dani}. 
Similar to~\cite{ye2022intrinsicnerf}, the normalized color loss calculated as $\|\operatorname{Nor}(\boldsymbol{A})-\operatorname{Nor}(\boldsymbol{I})\|$ is implemented to help Spin-UP learn a better albedo representation, where $\operatorname{Nor}(.)$ is the vector normalization. Following~\cite{li2023dani}, we calculate the boundary loss as the cosine similarity between the pre-computed and estimated boundary normal\footnote{Please refer to the supplementary material for more details about the setup of hyperparameters.}.

\begin{figure}[t]
    \centering
    \includegraphics[width=\linewidth]{"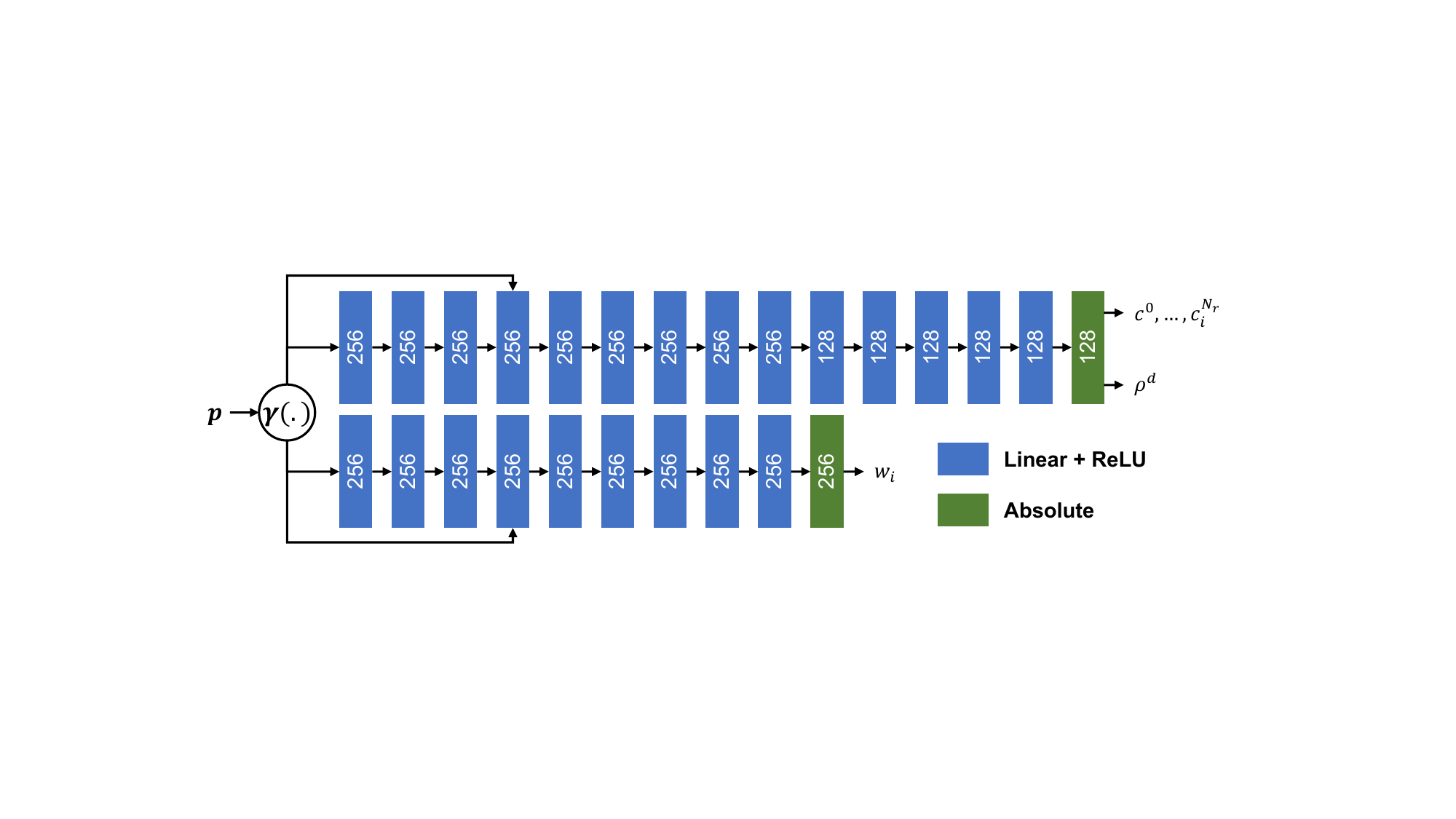"} 
    \caption{Proposed training strategies. (Top) \textbf{Interval sampling (IS)}. The high-resolution images are down-sampled into several low-resolution sub-images by extracting pixels with an interval of $N_B$, where $N_B=2$ in this example. (Bottom) \textbf{Shrinking range computing (SRC)}. Far points (yellow circles) that are $k$ ($k=3$ in this example) point away {\it w.r.t} the query position is selected to interpolate the close points (blue circles)'s depth for normal calculation. 
    During optimization, $k$ will be gradually reduced to 1.} 
    \label{fig:optimization}
    \vspace{-15pt}
\end{figure}

\subsection{Training Strategies}
\label{sec:ts}

Optimizing Spin-UP requires smoothness terms to facilitate convergence and avoid local optima. However, those terms are often implemented on full-resolution images, leading to high computational costs. To reduce those costs, we propose to use a sampling strategy noted as {\it interval sampling} (IS). To further improve convergence, we introduce another technique noted as {\it shrinking range computing} (SRC).

\noindent \textbf{Interval sampling (IS)}. IS samples ray batches from images to reduce the costs. Unlike random ray sampling~\cite{mildenhall2020nerf} or patch-based sampling~\cite{niemeyer2022regnerf}, IS preserves the object's shape. The idea is similar to downsampled techniques in~\cite{ikehata2023scalable,shi2016real}, but we don't merge the sub-images to full resolution. We experimentally find this strategy important for training on down-sampled sub-images with smoothness terms to avoid local optima (\Sref{sec:training strategies}). Specifically, we divide the image in full resolution into non-overlapping blocks, and each block contains $N_B\times N_B$ pixel points. By extracting pixel points from the same position in each block (\eg, the left-top pixel), we obtain $N_B\times N_B$ sub-images with a down-sampled resolution (See \Fref{fig:optimization} for illustration). During training, those sub-images are randomly sampled in each epoch, and the smoothness terms are calculated based on the sub-images resolution, which reduces the computational cost and ensures the effectiveness of those terms. 

\begin{table*}[t]
  \centering
  \caption{Qualitative comparison results in terms of MAE on shape group (left-top), light group (left-bottom), reflectance group (right-top), and spatially varying material group (right-bottom). \textbf{Bold} numbers indicate the best results in MAE. In light group, \{{\sc U., A., S., L.}\} stand for environment map named {\sc Urban, Attic, Studio}, and {\sc Landscape}, respectively. In the reflectance group, \{{\sc D. , S.}\} stand for material named {\sc Green Diff} and {\sc Green Spec}, respectively. In spatially varying material group, \{{\sc D. , S.}\} stand for material named {\sc Voronoi Diff} and {\sc Voronoi Spec}, respectively.}
  \resizebox{\linewidth}{!}{
    \begin{NiceTabular}{c|cccc|c|cccc|c}
    \hline
     \multirow{2}{*}{Method}& \multicolumn{5}{c}{Shape Group}       & \multicolumn{5}{c}{Reflectance Group} \\
     \cline{2-11}
          & \sc Ball  & \sc Bear  & \sc Buddha & \sc Reading & AVG   & \sc Pot2 (D.) & \sc Pot2 (S.) & \sc Reading (D.) & \sc Reading (S.) & AVG \\
          \midrule
    HY19~\cite{haefner2019variational}    & 41.32   & 53.88 & 67.90 & 54.85 & 54.49 & 57.45 & 37.43 & 65.48 & 58.04 & 54.60 \\
    S22~\cite{ikehata2022universal}   & 7.35  & 14.03 & 26.37 & 18.77  & 16.63  & 15.56  & 11.83  & 18.97  & 18.38  & 16.19  \\
    S23~\cite{ikehata2023scalable}   & 5.56  & 10.37 & 18.54 & 15.10  & 12.39  & 13.46  & 9.75  & 16.22  & 12.67  & 13.03  \\
    \midrule
    Spin-UP & \textbf{3.54} & \textbf{6.33} & \textbf{17.30} & \textbf{7.71} & \textbf{8.72} & \textbf{5.83} & \textbf{7.11} & \textbf{13.09} & \textbf{10.30} & \textbf{9.08} \\
    \midrule
    \midrule
     \multirow{2}{*}{Method}& \multicolumn{5}{c}{Light Group}       & \multicolumn{5}{c}{Spatially Varying Material Group} \\ \cline{2-11}
          & \sc Cow (U.) & \sc Cow (A.) & \sc Cow (S.) & \sc Cow (L.) & AVG   & \sc Pot2 (D.) & \sc Pot2 (S.) & \sc Reading (D.) & \sc Reading (S.) & AVG \\
          \midrule
    HY19~\cite{haefner2019variational} & 67.63 & 39.21 & 40.28 & 48.47 & 48.89 & 40.97 & 37.46 & 49.27 & 48.96 & 44.17 \\
    S22~\cite{ikehata2022universal}   & 17.17 & 12.74 & 17.11 & 11.35 & 14.59  & 18.59  & 17.63  & 22.80  & 23.75  & 20.69  \\
    S23~\cite{ikehata2023scalable}   & 11.93 & 7.52  & 12.38 & 11.60 & 10.84  & 14.22  & 11.00 & 14.58 & 14.31 & 13.53  \\
    \midrule
    Spin-UP  & \textbf{5.50} & \textbf{4.40} & \textbf{3.33} & \textbf{4.94} & \textbf{4.54} & \textbf{5.58} & \textbf{6.97} & \textbf{12.54} & \textbf{11.52}  & \textbf{9.15} \\
    \hline
    \end{NiceTabular}%
    }
    \vspace{-10pt}
  \label{tab:syn}
\end{table*}%

\noindent \textbf{Shrinking range computing (SRC)}. Without merging sub-images back to a full-resolution image in IS, there will be an aliasing issue in the inverse rendering process. Such an issue is caused by the fact that the normal calculation in our framework on sub-images requires four adjacent points' depths in sub-images resolution~\cite{li2023dani}, which degrades the precision of normal calculation\footnote{According to normal's definition, the smaller the distance between the adjacent points and the query points (red circle in \Fref{fig:optimization}), the more accurately representing the geometry at query points. Therefore, the blue circles' depths are preferred for normal calculation. }. Therefore, SRC is applied for anti-aliasing. It uses points adjacent to the query point (blue circles in \Fref{fig:optimization}) {\it in the full-resolution image coordinates} to calculate the normal for each pixel in the sub-images. Such a strategy maintains the precision of normal calculation.  However, at the early stage of training, calculating the normal based on the blue circles' depth is vulnerable to perturbation in per-pixel training. Therefore, SRC gradually selects points (yellow circles) from far ($k=3$ points away) to close (blue circles) {\it w.r.t} query points to interpolate the blue circles' depth, as normal calculation on far points' depth will lead to a more smooth and stable normal map at the early stage, which eventually improves convergence, as validated in \Sref{sec:training strategies}.

\section{Experiments}

We validate the effectiveness of the proposed Spin-UP on synthetic and real-world data. We use mean angle error (MAE) to evaluate the normal map's reconstructed quality and PU-PSNR~\cite{azimi2021pu21}, and PU-SSIM~\cite{azimi2021pu21} to evaluate the reconstructed environment light. Since no existing datasets follow our spin light setup, we collect the dataset using Blender and our device. 

\subsection{Evaluation on Synthetic Datasets}
\label{sec:eval synth}
We collect several objects, environment maps, and materials to render the synthetic dataset in Blender by Cycles. Specifically, five shapes from the DiLiGenT-MV dataset~\cite{li2020multi} (\ie, {\sc Buddha}, {\sc Bear}, {\sc Cow}, {\sc Pot2}, and {\sc Reading}) and a generated shape ({\sc Ball}), five HDR environment maps (\ie, {\sc Landscape}, {\sc Quarry}, {\sc Urban}, {\sc Attic}, {\sc Studio}), two PBR materials (\ie, {\sc Rusty Steel}, {\sc Leather}), and four synthetic materials (\ie, {\sc Voronoi Diff, Voronoi Spec, Green Diff, Green Spec})\footnote{The generated pattern for {\sc Voronoi Diff} and {\sc Voronoi Spec} follow similar setup in CNN-PS~\cite{ikehata2018cnn}.} are used for evaluation.  
We devise four groups of data for evaluation: shape group, light group, reflectance group, and spatially varying material group, each containing four scenes.
For each scene, 50 observed images with a resolution of $512 \times 512$ are rendered by a perspective camera with the focal of $50 \mathrm{mm}$ and a frame size of $36 \mathrm{mm}\times 36 \mathrm{mm}$. 
The camera rotation $\theta$ for consecutive images follows a non-uniform rotation velocity. We compare Spin-UP with three advanced NaUPS methods, including two supervised NaUPS methods (S22~\cite{ikehata2022universal} and S23~\cite{ikehata2023scalable}) and one unsupervised UPS method (HY19~\cite{haefner2019variational})\footnote{Please refer to the supplementary material for all the qualitative comparison between Spin-UP and other methods on the synthetic and real-world dataset.}.

\noindent\textbf{Normal estimation comparison}. According to results in \Tref{tab:syn}, Spin-UP presents a superior performance compared to all other NaUPS methods. Specifically, in {\it shape group}, the low MAE on {\sc Ball, Bear, Buddha, Reading} in {\sc Rusty Steel} rendered under {\sc Quarry} indicate the practicability of Spin-UP to various shapes.
In {\it light group}, the low-variance of MAE ($0.81^\circ$ for Spin-UP vs. $1.94^\circ$ for S23~\cite{ikehata2023scalable}) on {\sc Cow} in {\sc Leather} rendered under {\sc Quarry, Urban, Attic,} and {\sc Studio} demonstrates robustness toward different environment lights. In {\it reflectance group}, the results on {\sc Pot2} and {\sc Reading} in {\sc Green Diff} and {\sc Green Spec} rendered under {\sc Landscape} demonstrate the ability to handle non-Lambertian objects. 
In {\it spatially varying material group}, results on {\sc Pot2} and {\sc Reading} rendered in {\sc Voronoi Diff} or {\sc Voronoi Spec} under {\sc Landscape} prove adaptability to challenging scenarios. 
A comparative analysis of the outcomes of the reflectance group and the spatially varying material group in our method reveals that the MAE remains relatively consistent across identical objects with different materials, underscoring the robustness of Spin-UP in handling diverse materials. Also, we find that Spin-UP sometimes performs better on specular objects than on diffuse objects (\ie, {\sc Reading}). We attribute this to the high-frequency details in specular reflectance that may be useful for shape-light reconstruction.



\begin{figure}[t]
    \centering
    \includegraphics[width=\linewidth]{"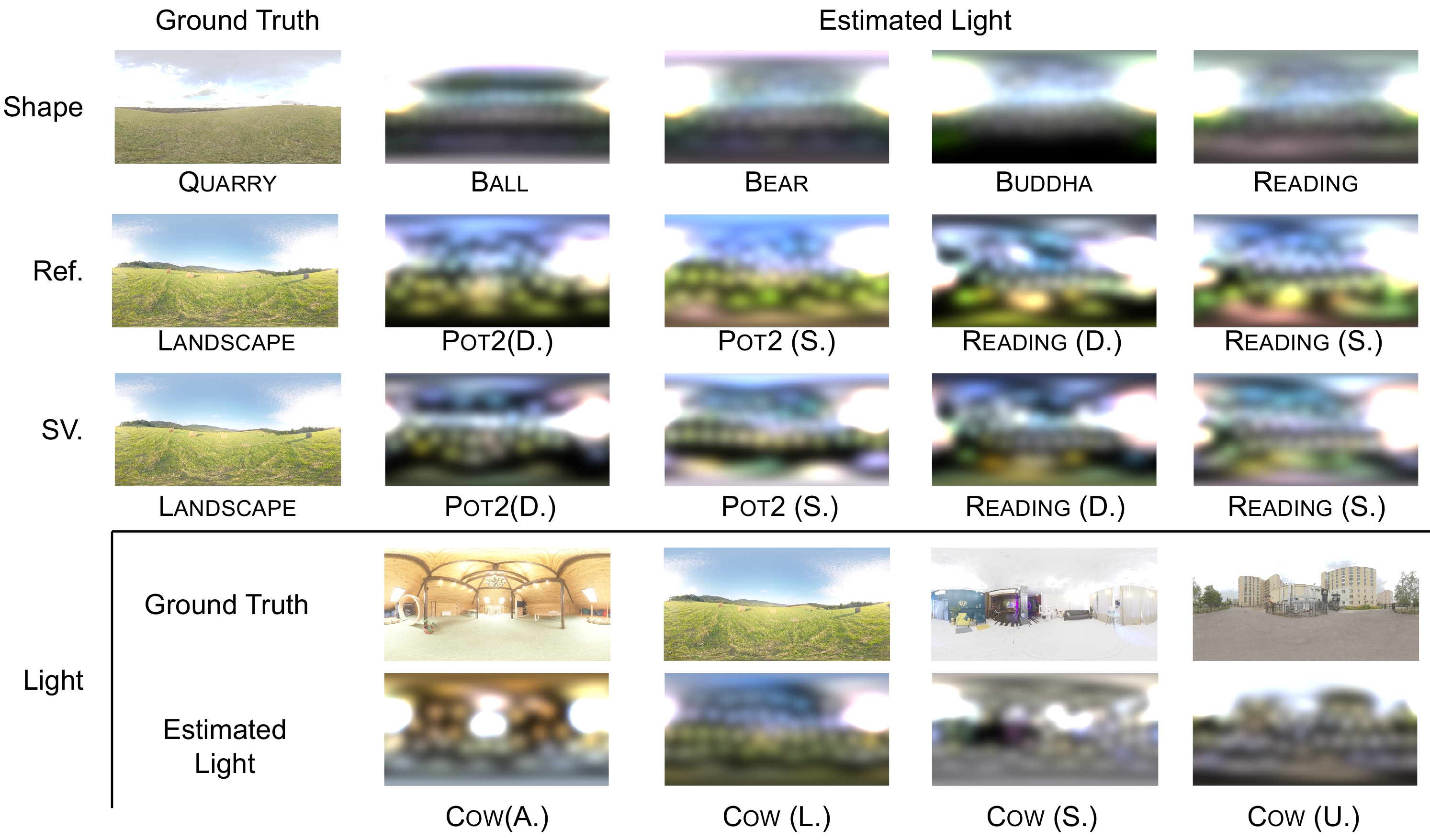"}
    \caption{The visual quality comparison of light between the estimated one by Spin-UP (columns 2-5) and the ground truth (column 1) on the four groups, \ie, shape (row 1), reflectance (row 2), spatially varying material (row 3), and light group (row 4-5). The intensity of the estimated light maps is scaled for visualization.}
    \label{fig:light_est}
    \vspace{-15pt}
\end{figure}

\begin{figure}[t]
    \centering
    \includegraphics[width=\linewidth]{"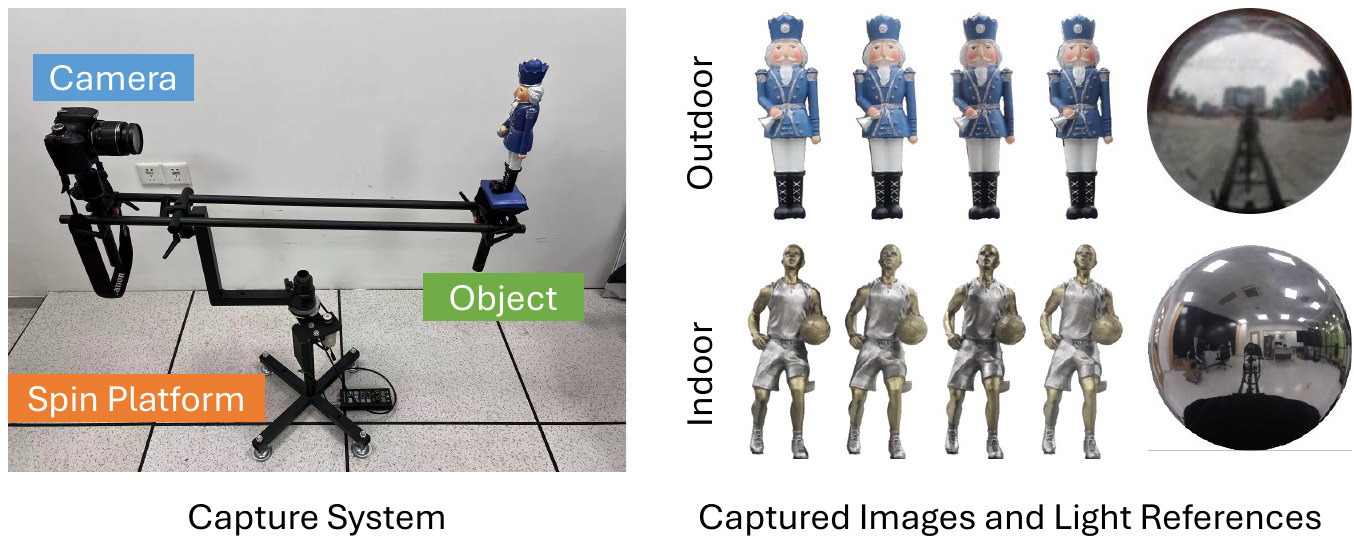"} 
    \caption{Illustration of the device for real data collection. Please refer to the supplementary material for more details. Left: the capture system contains a camera, a spin platform, and an object. Right: Captured images and paired mirror balls (as light reference) of two objects in the indoor and outdoor scenes, respectively.} 
    \vspace{-10pt}
    \label{fig:data_collection}
\end{figure}

\noindent \textbf{Light estimation comparison}. 
\Fref{fig:light_est} provides a qualitative comparison between the estimated environment light and the ground truth on four groups. 
We can observe that the learned light map reflects the position of the light source, especially in {\sc Cow (S.)} given such a challenging setup without any prior information about the object material or shape.
It also reconstructs a reasonable light map for objects with diffuse reflectance, such as {\sc Pot2 (D.)} and {\sc Reading (D.)}, 
further highlighting the effectiveness of the proposed light initialization. 
However, it should be noted that the estimated environment light is influenced by the object material, particularly when handling objects with a nearly uniform base color (\eg, reflectance group) or spatially varying material, which may generate unpleasant artifacts (\eg, inconsistent color in estimated environment light of {\sc Reading (D.)} in the spatially varying material group). Those artifacts are hard to eliminate without priors.

\subsection{Evaluation on Real-world Datasets}
\begin{figure}[t]
    \centering
    \includegraphics[width=\linewidth]{"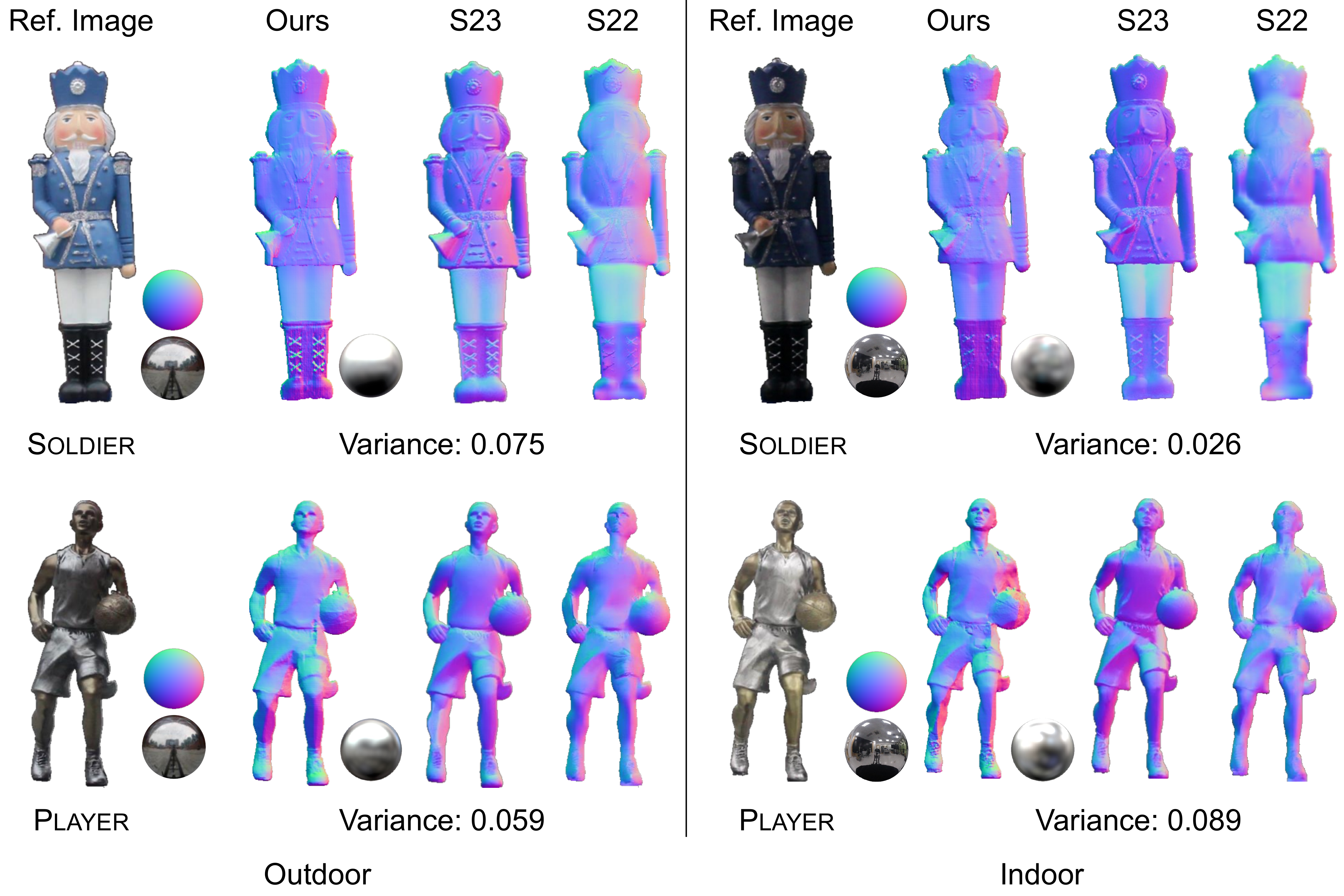"} 
    \vspace{-10pt}
    \caption{Qualitative comparison for the estimated normal map and environment light on {\sc Soldier} and {\sc Player} capturing under outdoor (columns 1-6) and indoor (columns 7-12) between ours (column 3, 9), S23~\cite{ikehata2023scalable} (column 5, 11), and S22~\cite{ikehata2022universal} (column 6, 12). We also show the reference sphere and reference environment light (columns 2, 7), and estimated light (columns 4, 10).} 
    \vspace{-15pt}
    \label{fig:real-world}
\end{figure}

We set up our spin light capture system to collect real data under indoor and outdoor scenes, as shown in \Fref{fig:data_collection}.
After preprocessing, we end up with 50 images for each object, with a $540 \times 540$ resolution. Here, we showcase the normal estimation results on two objects (\ie, {\sc Soldier}, and {\sc Player}) under indoor and outdoor environment lights in~\Fref{fig:real-world}, compared with S23~\cite{ikehata2023scalable}, S22~\cite{ikehata2022universal}. 
We do not show HY19~\cite{haefner2019variational}'s result as it failed on the captured data.

\noindent \textbf{Normal estimation comparison}. According to the \Fref{fig:real-world}, Spin-UP has competitive performance compared to the state-of-the-art supervised method~\cite{ikehata2023scalable}. 
In some scenarios, we recover more reasonable results regarding the overall normal distribution than the reference sphere, particularly for {\sc Soldier} indoor and {\sc Player} indoor. 
Spin-UP can effectively capture high-frequency details such as wrinkles on clothes in {\sc Player} and {\sc Soldier}. 
By contrast, S23~\cite{ikehata2023scalable} may contain artifacts (\ie, an incorrect normal map distribution) even though they have more details than ours, and S22~\cite{ikehata2022universal} generates over-smooth results. 
By comparing indoor and outdoor results, we observe that the performance of S23~\cite{ikehata2023scalable} degrades significantly in indoor scenarios, which may be attributed to the data bias and low pixel variance, while our method is not greatly affected. 




\section{Ablation Study}
\label{sec:abl}

\subsection{Light Initialization Validation}
To comprehensively validate the effectiveness of our light initialization method, we conduct experiments in two aspects: a comparison of different light initialization methods and the effectiveness of the filters.

\begin{table}[t]
\setlength{\tabcolsep}{15pt}
  \centering
  \caption{Ablation studies on Spin-UP's alternatives regarding average PU-PSNR~\cite{azimi2021pu21} and PU-SSIM~\cite{azimi2021pu21} on synthetic dataset. `Fib.' and `Rand.' represent the Fibonacci and random initialization method, respectively. \textbf{Bold} number indicates the best results.}
  \vspace{-3pt}
  \resizebox{0.8\linewidth}{!}{
    \begin{tabular}{lccc}
    \hline
          & {\it w} Rand. & {\it w} Fib. & Spin-UP\\
    \midrule
    PU-PSNR~\cite{azimi2021pu21}$\uparrow$ & 16.86  & 18.98 & \textbf{22.06} \\
    PU-SSIM~\cite{azimi2021pu21}$\uparrow$ & 0.45 & 0.52 & \textbf{0.60}\\
    \hline
    \end{tabular}%
  \label{tab:lgt_quant}%
  }
      \vspace{-15pt}
\end{table}%

\noindent \textbf{Comparison on light initialization methods}. 
We compare our light initialization method with two widely used SG light initialization methods, \ie, the random initialization noted as `{\it w} Rand.' and Fibonacci lattice~\cite{zhang2021physg} noted as `{\it w} Fib.'. A quantitative comparison of reconstructed normal and light maps is shown in \Tref{tab:lgt_quant} and \Tref{tab:abl}. Compared with `{\it w} Rand.', the improvement in average ($1.75^\circ$ reduction in MAE on normal estimation and $5.20$ / $0.15$ increase in PU-PSNR / PU-SSIM on light estimation, respectively) indicates our method's adaptability. Compared with `{\it w} Fib.', we observe an obvious advantage in the shape group ($0.61^\circ$ reduction in MAE), while a smaller advantage in the spatially varying material group ($0.19^\circ$ reduction in MAE). This is because the material and shape will affect the initial environment light's quality. While the estimated environment light is most accurate on smooth geometry with simple material, the quality will degrade on complex geometry and spatially varying material.



\noindent \textbf{Comparison on the designed filters}. We compare Spin-UP with three alternatives (\ie, {\it w/o} $\mathcal{F}^c$, {\it w/o} $\mathcal{F}^d_{SH}$, and {\it w/o} $\mathcal{F}^d_{TH}$). The results in~\Tref{tab:abl} demonstrate the effectiveness of those filters. 
Specifically, dropping $\mathcal{F}^d_{TH}$ will lead to mismatching light source position in the initialized environment light introduced by the specular reflectance, which eventually affects the accuracy of the estimated normal; dropping $\mathcal{F}^d_{SH}$ will harm the performance, especially in the spatially varying material group ($6.58^{\circ}$ increase in MAE) since $\mathcal{F}^d_{SH}$ is essential in extracting low-frequency reflectance to initialize the environment light; dropping $\mathcal{F}^c$ will increase MAE in average ($0.32^{\circ}$), illustrating the necessity of reducing chromatic bias. 

\vspace{-3pt}
\subsection{Training Strategies Validation}
\label{sec:training strategies}
\vspace{-3pt}
The interval sampling will facilitate the training of Spin-UP in two ways. 
First, the training time is two times shorter (25 min per object on average vs. 60 min on average, depending on the image's valid points for the object), and the GPU memory occupation is five times smaller (around 5 GB vs. 25 GB during training) than directly training on full-image resolution. 
Second, comparing Spin-UP with `{\it w/o} Intv.', which applies a random sampling strategy and calculates the smoothness terms on patches ($3 \times 3$ pixels), we find that the performance drops $0.52^\circ$ on average, and most ($0.97^\circ$) on the spatially varying material group. 
This is because the patch-based smoothness may not work uniformly on different parts of the object, diminishing the effectiveness of smoothness terms, especially on objects in spatially varying material with abrupt texture changes. 
\begin{table}[t]
\setlength{\tabcolsep}{12pt}
  \centering
  \caption{Ablation studies on Spin-UP's alternatives regarding average MAE on four groups (shape, light, reflectance, and spatially varying material group). `Fib.', `Rand.', `Intv.', and `Shrk.' represent the Fibonacci initialization method, random initialization, interval sampling, and shrinking range computing, respectively. \textbf{Bold} number indicates the best results in MAE.}
  \vspace{-3pt}
  \resizebox{1.\linewidth}{!}{
    \begin{threeparttable}
    \begin{tabular}{lcccc|c}
    \hline
          & Shape & Light & Ref. & SV. & AVG \\
    \hline
    {\it w} Rand. & 12.02  & 6.44 & 9.79  & 10.28 & 9.60 \\
    {\it w} Fib. & 9.33 & 5.34 & 10.20 & 9.34 & 8.55 \\
    {\it w/o} Intv. & 8.87 & 5.02 & 9.83 & 10.12 & 8.37 \\
    {\it w/o} Shrk. & 10.10 & 4.92 & 10.12 & 10.43 & 8.81 \\
    {\it w/o} $\mathcal{F}^c$ & 9.44  & 4.61  & 9.24 & 9.40 & 8.17 \\
    {\it w/o} $\mathcal{F}^d_{SH}$ & 9.75  & 8.29 & \textbf{8.93}  & 15.73 & 10.41 \\
    {\it w/o} $\mathcal{F}^d_{TH}$ & 9.30  & 5.04  & 9.18  & 12.49 & 8.82 \\
    Spin-UP & \textbf{8.72}  & \textbf{4.54}  & 9.08  & \textbf{9.15} & \textbf{7.85} \\
    \hline
    S23~\cite{ikehata2023scalable}\dag & 12.42  & 8.56 & 12.52  & 12.33 & 11.46 \\
    Spin-UP\dag & 11.62  & 9.25 & 11.07  & 9.07 & 9.48 \\
    \hline
    \end{tabular}%
    \begin{tablenotes}
    \footnotesize
    {\item[\dag] Method with \dag~is tested on dataset with point light + environment light~\Sref{sec:pl}.}
    \end{tablenotes}
    \end{threeparttable}
  \label{tab:abl}%
  }
      \vspace{-15pt}
\end{table}%
The shrinking range computing helps avoid local optima when training Spin-UP on down-sampled images while still using full-resolution image coordinates for normal calculation. We compared Spin-UP with the alternative `{\it w/o} Shrk.', which does not implement this strategy. 
The average MAE on normal estimation for four groups increases $0.96^\circ$, highlighting the importance of this strategy. 

\subsection{Additional Validation on Point Light Source}
\label{sec:pl}
\begin{figure}[t]
    \centering
    \includegraphics[width=0.47\textwidth]{"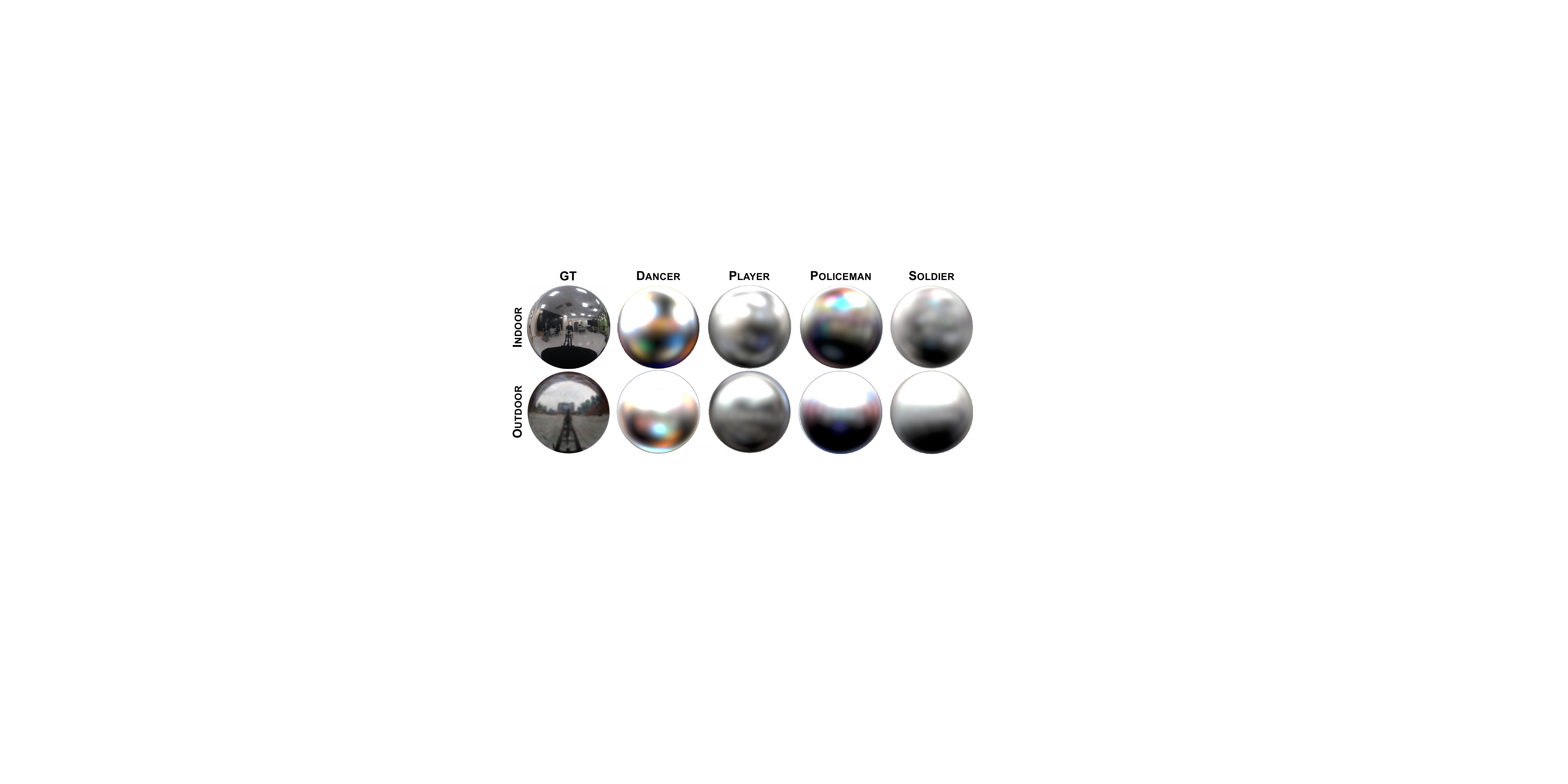"} 
    \caption{(a) Illustration of new light setup. (b) Qualitative comparison on {\sc Soldier} between our method and S23~\cite{ikehata2023scalable} with dominant point light.} 
    \label{fig:pl}
    \vspace{-15pt}
\end{figure}

To ensure a more fair comparison with the state-of-the-art supervised methods (S23~\cite{ikehata2023scalable}), we add a dominant point light to the environment light in synthetic and real-world dataset\footnote{Point light's setup follows~\cite{ikehata2023scalable} and \cite{liu2018near}.}. According to~\Tref{tab:abl}, the proposed Spin-UP has a lower MAE on estimated normal maps than S23~\cite{ikehata2023scalable} on the synthetic dataset ($9.48^\circ$ ours v.s. $11.46^\circ$ S23~\cite{ikehata2023scalable}). As shown in \Fref{fig:pl}, we have a visually comparable result on the real-world dataset given far-field point light ($2\text{m}$) and a better result given near-field point light ($0.4\text{m}$), validating the adaptability of Spin-UP on unseen light sources.

\vspace{-2pt}
\section{Conclusion}
\vspace{-2pt}
This paper proposes Spin-UP to address NaUPS in an unsupervised manner. Thanks to our setup to mitigate the ill-posedness, the light initialization method to alleviate the ambiguities of NaUPS, and the proposed training strategies to facilitate fast convergence, Spin-UP can recover surfaces with isotropic reflectance under various lights. Experiments in synthetic and real-world datasets have shown that Spin-UP is robust to various shapes, lights, and reflectances.

\noindent\textbf{Limitations and future work.} Although Spin-UP is efficient and robust in solving NaUPS, it has several limitations: 1) Spin-UP assumes infinitely far light sources, which omit the spatially varying lighting; 2) the materials' base color will bias the estimated environment light; 3) Spin-UP assumes objects to have isotropic reflectance, ignoring inter-reflections and anisotropic features, meaning that it cannot perform well on objects with anisotropic reflectance, such as aluminum, or strong inter-reflections, such as a glass bowl; 
4) Spin-UP doesn't compute the shadow iteratively, which may result in artifacts on objects with complicated shapes. Overcoming the above limitations will be regarded as our future work. Besides, the rotation axis of our device does not align with the object's center due to the consideration of structural stability, which may introduce bias from the spatially varying light in observed images. Redesigning the image capture device is also one of the future works. At last, we find it interesting to improve the setup by relieving the requirement for single-axis $360^\circ$ rotation to free rotations for easier implementation on portable devices.

\noindent \textbf{Acknowledgments.} This work is supported by the National Natural Science Foundation of China under Grant No. 62136001, 62088102, Rapid-Rich Object Search (ROSE) Lab, Interdisciplinary Graduate Programme, Nanyang Technological University, Singapore, and the State Key Lab of Brain-Machine Intelligence, Zhejiang University, Hangzhou, China.

\clearpage

{
    \small
    \bibliographystyle{ieeenat_fullname}
    \bibliography{main}
}

\onecolumn
\appendix
\section*{Supplementary Material}

\noindent In this supplementary material, 
\begin{enumerate}
    \item we give more implementation details in~\Sref{sec:implement}, including details of framework structure (footnote 6) and hyperparameters setup (footnote 7). 
    \item we introduce more about boundary normal calculation and normal calculation for rendering equation in perspective projection in \Sref{sec:bound} (footnote 6);
    \item we provide an overview of the synthetic and real-world dataset in \Sref{sec:dataset}. We also explain how we collect and preprocess the real-world dataset;
    \item we showcase a qualitative comparison between Spin-UP and other methods on the real-world dataset in \Sref{sec:qualitative comparison} (footnote 10). More results from the real-world dataset are also included in this section (footnote 10);
\end{enumerate}

\section{Implementation Details}
\label{sec:implement}
\subsection{Network Structure}
\begin{figure}[h]
    \vspace{-5pt}
    \centering
    \includegraphics[width=0.7\textwidth]{"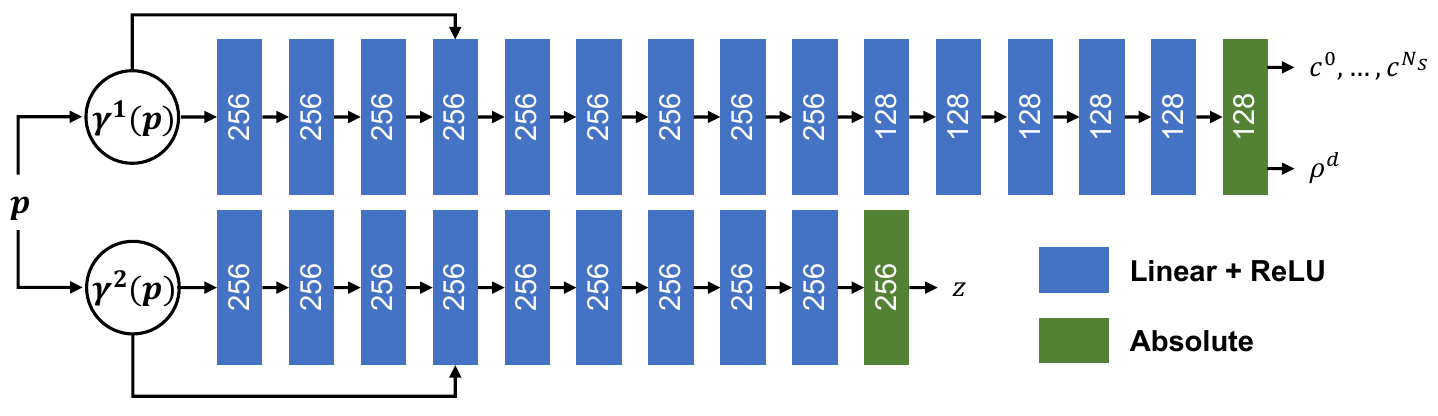"}
    \caption{Network structure of MLPs for depth and material estimation in Spin-UP.}
    \label{fig:struct}
    \vspace{-10pt}
\end{figure}

We use the similar multi-layer perceptrons (MLPs)' structures in~\cite{li2022self, li2023dani}, shown in~\Fref{fig:struct}. The input of MLPs is pixels' 2D coordinate ($p=(x, y)$) in an image, which will pass through a positional encoding module similar in~\cite{mildenhall2020nerf} calculated as
{\small
\begin{align}
    \begin{aligned}
        \gamma(p)= \left(\sin \left(2^0 \pi p\right), \cos \left(2^0 \pi p\right), \cdots, \sin \left(2^{L_p-1} \pi p\right), \cos \left(2^{L_p-1} \pi p\right)\right),
    \end{aligned}
\end{align}}
where $L_p$ is the positional code's dimension, set as 10 for $\gamma^1(.)$ and 6 for $\gamma^2(.)$

\subsection{Loss Functions and Hyperparameters Setup}
In Spin-UP, we implement:
\begin{enumerate}
    \item L1 inverse rendering loss $L_{r}$ calculated as,
    $\sum_{i=1}^{N_P} \sum_{j=1}^{N_I} |\boldsymbol{m}_{i j} - \hat{\boldsymbol{m}}_{i j}|$.
    \item Normalized color loss $L_{\mathrm{color}}$, calculated as,  $\lambda_c \|\mathrm{Nor}(\boldsymbol{A}) - \mathrm{Nor}(\boldsymbol{I})\|$, where $\lambda_c=0.5$.
    \item Boundary loss $L_{\mathrm{b}}$, calculated as the cosine similarity between the pre-computed and estimated boundary normal.
    \item Smoothness terms $L_{\mathrm{sm}}$ on albedo map $\boldsymbol{A}$, normal map $\boldsymbol{N}$, spatially varying Gaussian bases weights $c^n$, is calculated as, 
    \begin{align}
        \begin{aligned}
        L_{\mathrm{sm}}
        =\frac{\lambda}{N_P} \sum_{i=1}^{N_P}\left|\frac{\partial \boldsymbol{A}}{\partial x} +\frac{\partial \boldsymbol{A}}{\partial y}\right|
        +\frac{\lambda_N}{N_P} \sum_{i=1}^{N_P}\left|\frac{\partial \boldsymbol{N}}{\partial x}
        +\frac{\partial \boldsymbol{N}}{\partial y}\right|
        + \frac{\lambda_S}{N_P} \sum_{n=1}^{N_S}\sum_{i=1}^{N_P} \left|\frac{\partial c^n_i}{\partial x} + \frac{\partial c^n_i}{\partial y}\right|,
        \end{aligned}
    \end{align}
    where, $\lambda = 0.01$, $\lambda_N = 0.02$, $\lambda_S = 0.01$. 
\end{enumerate}

\noindent We train the Spin-UP in three stages similar to~\cite{li2023dani}. For the first stage, the loss $\mathcal{L}_\mathrm{stage1}$ is calculated as below for a faster convergence.
\begin{align}
    \begin{aligned}
        \mathcal{L}_\mathrm{stage1} = \mathcal{L}_r + \mathcal{L}_{\mathrm{b}} + \lambda_c \mathcal{L}_\mathrm{color} + \mathcal{L}_\mathrm{sm},
    \end{aligned}
\end{align}
For the second stage, we drop the smoothness term on the albedo map and reduce $\lambda_N$ to 0.05 for details refinement, where $\mathcal{L}_N=\operatorname{TV}(\boldsymbol{N})$ is the smoothness term on normal map.
\begin{align}
    \begin{aligned}
        \mathcal{L}_\mathrm{stage2} = \mathcal{L}_r + \mathcal{L}_{\mathrm{b}} + \lambda_c \mathcal{L}_\mathrm{color} + \lambda_{N
        } \mathcal{L}_N,
    \end{aligned}
\end{align}
For the third stage, we drop the smoothness terms $\mathcal{L}_N$ to further refine the details.
\begin{align}
    \begin{aligned}
        \mathcal{L}_\mathrm{stage3} = \mathcal{L}_r + \mathcal{L}_{\mathrm{b}} + \lambda_c \mathcal{L}_\mathrm{color}.
    \end{aligned}
\end{align}
The three stages take 500, 1000, and 500 epochs, respectively. During training, we use Adam as the optimizer with a learning rate $\alpha=0.001$ and a batch size of 4 images per iteration.

\section{Normal Calculation in Perspective View}
\label{sec:bound}
\subsection{Boundary Normal Calculation}
\begin{minipage}{0.9\linewidth}
      \centering
      \begin{minipage}{0.45\linewidth}
          \begin{figure}[H]
              \centering
              \includegraphics[width=\linewidth]{"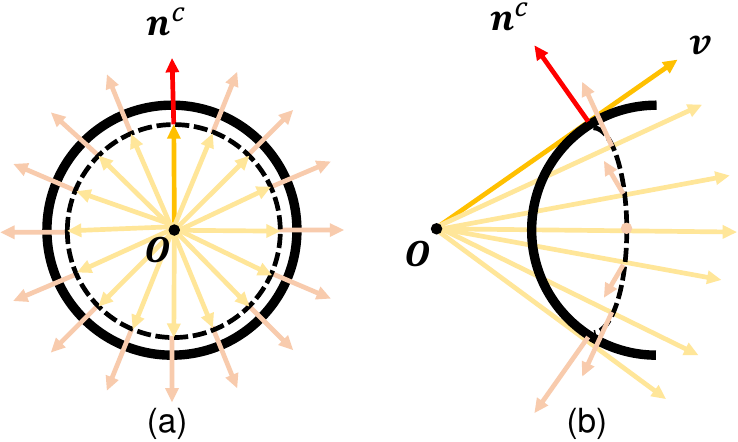"}
              \caption{An illustration of occluding boundaries' normal relationship with view directions for (a) front view and (b) side view of a surface. The dotted line in (b) indicates the outermost boundaries of an object in perspective projection.}
              \label{fig:perspective}
          \end{figure}
      \end{minipage}
      \hspace{0.05\linewidth}
      \begin{minipage}{0.45\linewidth}
          \begin{figure}[H]
              \centering
              \includegraphics[width=\linewidth]{"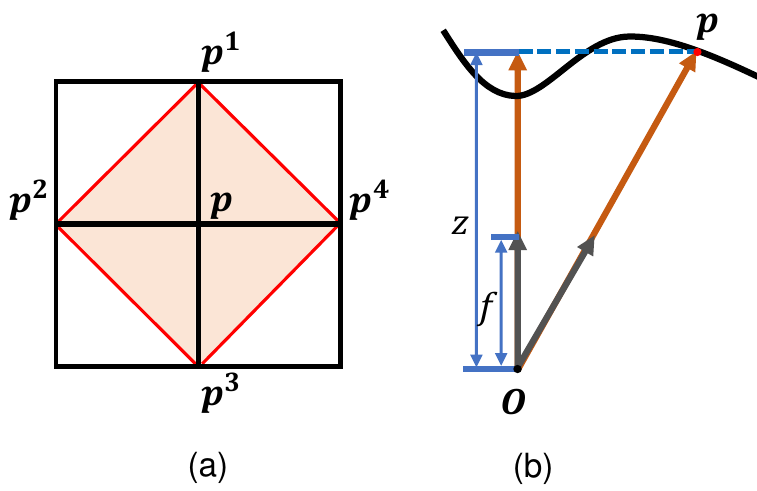"}
              \caption{An illustration of (a) adjacent points' positions for normal fitting method~\cite{li2023dani} in perspective projection, (b) \Eref{eq:persp}. }
              \label{fig:nml_fit}
          \end{figure}
      \end{minipage}
  \end{minipage}
  
In perspective projection, the surface normal is perpendicular to the object's occluding boundaries $B(x, y)$ and view direction $\boldsymbol{v}$, as shown in \Fref{fig:perspective}. Therefore, the boundaries' normal $\boldsymbol{n}^b$ is calculated as
\begin{align}
    \begin{aligned}
        \boldsymbol{n}^b \cdot \boldsymbol{v}^b = 0, ~\boldsymbol{n}^b \cdot (\frac{\partial B}{x}, \frac{\partial B}{y}, 1)^\top= 0, 
    \end{aligned}
\end{align}
In practice, the outer boundaries of an object in images may not precisely match its actual boundaries due to limited image resolution. Therefore, we add a small offset ($\beta=0.1$) to make the pre-computed boundaries normal more accurate:
\begin{align}
    \begin{aligned}
        \boldsymbol{n}^b = \mathrm{Nor}(n^{bx}, n^{by}, n^{bz} + \beta).
    \end{aligned}
\end{align}

\subsection{Normal Calculation For Rendering Equation}
The normal fitting method \cite{li2023dani} in orthogonal projection is shown below:
\begin{align}
    \begin{aligned}
        \boldsymbol{n} & =\sum_{k=1}^4 \gamma^k \boldsymbol{n}^k=\sum_{k=1}^4 \gamma^k \operatorname{Nor}\left[\left(\boldsymbol{p}^{k+1}-\boldsymbol{p}\right) \times\left(\boldsymbol{p}^k-\boldsymbol{p}\right)\right]^{\top}, \\
        \gamma^k & =\frac{\left|d^k\right|^{-1}}{\sum_{k=1}^4\left|d^k\right|^{-1}}, \quad d^k=z^k+z^{k+1}-2 z,
    \end{aligned}
    \label{eq:nml_fit}
\end{align}
where, $\boldsymbol{p}^k=(x^k, y^k, z^k)$ is the adjacent point of the query point $\boldsymbol{p}=(x, y, z)$ and $x, y \in [-1, 1]$, $k = 1$ if $k + 1 > 4$, as shown in \Fref{fig:nml_fit}, (a). To extend the normal fitting method to the perspective projection, we first compute the points' coordinates in the camera coordinate system by
\begin{align}
    \begin{aligned}
        \boldsymbol{p}^{k\prime} & =(x^k\frac{z^k}{f}s_x, y^k\frac{z^k}{f}s_y, z^k), \\
        \boldsymbol{p}^{\prime} & =(x\frac{z}{f}s_x, y\frac{z}{f}s_y, z).
    \end{aligned}
    \label{eq:persp}
\end{align}
where $f$ is the camera's focal, $s_x$ and $s_y$ are half of the width and height of the camera's frame. Replace $\boldsymbol{p}^{k}$ and $\boldsymbol{p}$ in \Eref{eq:nml_fit} by $\boldsymbol{p}^{k\prime}$ and $\boldsymbol{p}^{\prime}$, we get the normal fitting method in perspective projection.

\section{Datasets}
\label{sec:dataset}
\subsection{Synthetic Dataset}
\begin{figure}[t]
    \centering
    \includegraphics[width=.9\textwidth]{"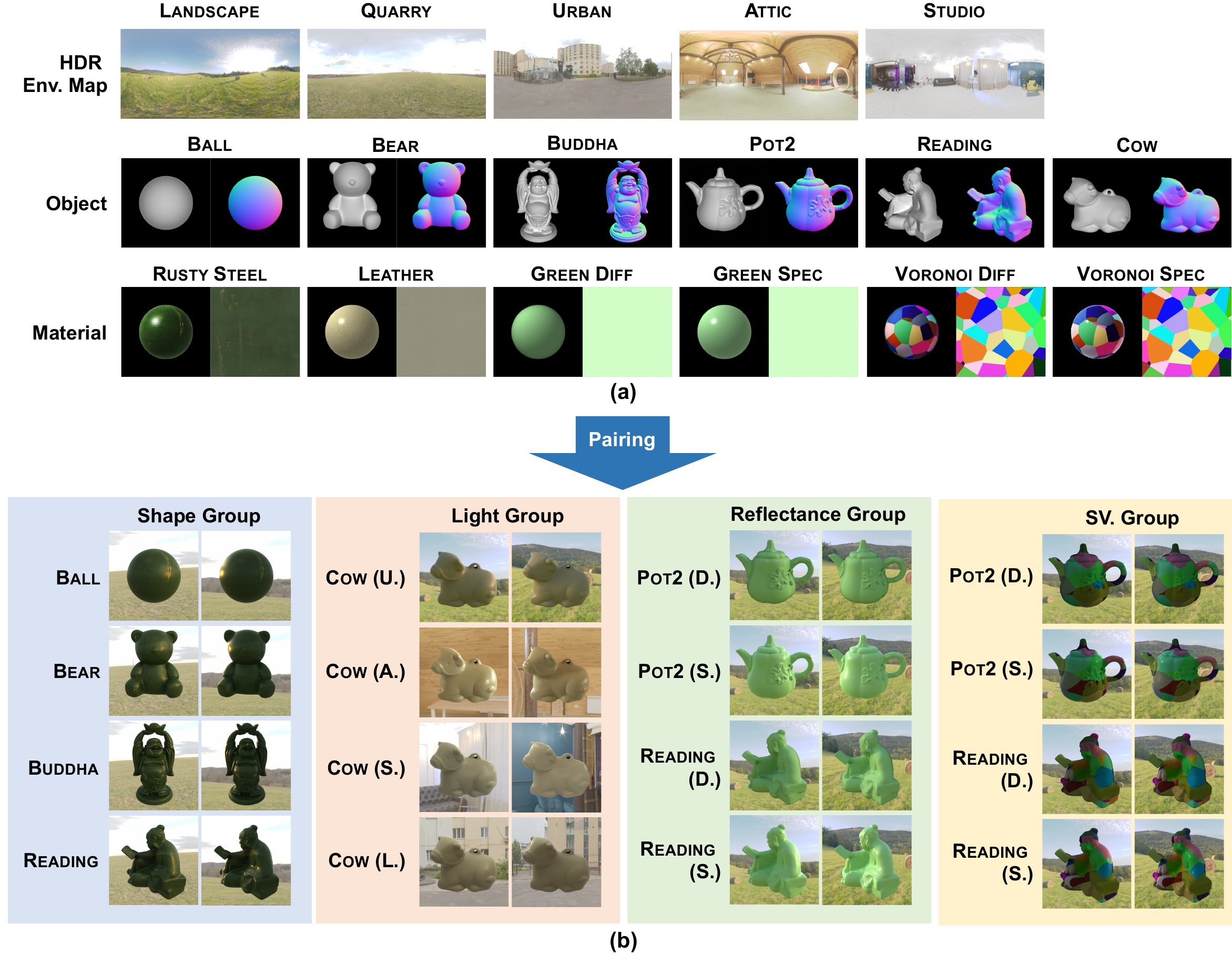"}
    \caption{(a) HDR environment maps (row 1), objects (row 2), and materials (row 3) involved in the synthetic dataset. Each figure in row 2 consists of two subfigures for 3D model preview (left) and normal map (right). Each figure in row 3 consists of two subfigures for material rendered on a sphere (left) and albedo (right). (b) Example images from each scene in four groups.}
    \label{fig:syn}
\end{figure}

In \Fref{fig:syn}, we showcase all 5 objects with 6 materials under 5 HDR environment maps rendered by Blender Cycles\footnote{\url{https://www.blender.org}}. This results in 16 scenes\footnote{One scene representing an object with one material rendered under HDR environment maps.} of synthetic data that are classified into 4 groups, \ie, the shape group, light group, reflectance group, and spatially varying group. 

\begin{table}[t]
\caption{Length, width, height, and capturing distance for {\sc Soldier}, {\sc Player}, {\sc Dancer}, {\sc Policeman} and {\sc Eevee}.}
    \label{tab: object}
    \centering
    \vspace{0pt}
    \resizebox{0.6\linewidth}{!}{
    \begin{NiceTabular}{c|ccccc}
    \hline
    Properties & {\sc Soldier} & {\sc Player}  & {\sc Dancer} & {\sc Policeman} & {\sc Eevee}\\
    \midrule
    Length (cm)  & 9.50 & 11.50 & 4.00 & 4.00 & 4.00 \\
    Width (cm)   & 7.00 & 11.00 & 5.00 & 4.00 & 4.00\\
    Height (cm)   & 3.00 & 28.00 & 4.00 & 9.00 & 9.00\\
    Distance (m) & 0.90 & 0.90  & 0.40 & 0.40 & 0.30\\
    \hline
    \end{NiceTabular}
    }
    \vspace{-15pt}
\end{table}

\subsection{Real-world Dataset}
The real-world dataset contains 5 objects captured under indoor and outdoor environments with spatially-varying materials. The five real-world objects used in our study are the {\sc Soldier}, {\sc Player}, {\sc Policeman}, {\sc Dancer}, and {\sc Eevee}. The objects' sizes are shown in \Tref{tab: object}.

\begin{figure}[t]
    \centering
    \includegraphics[width=0.7\textwidth]{"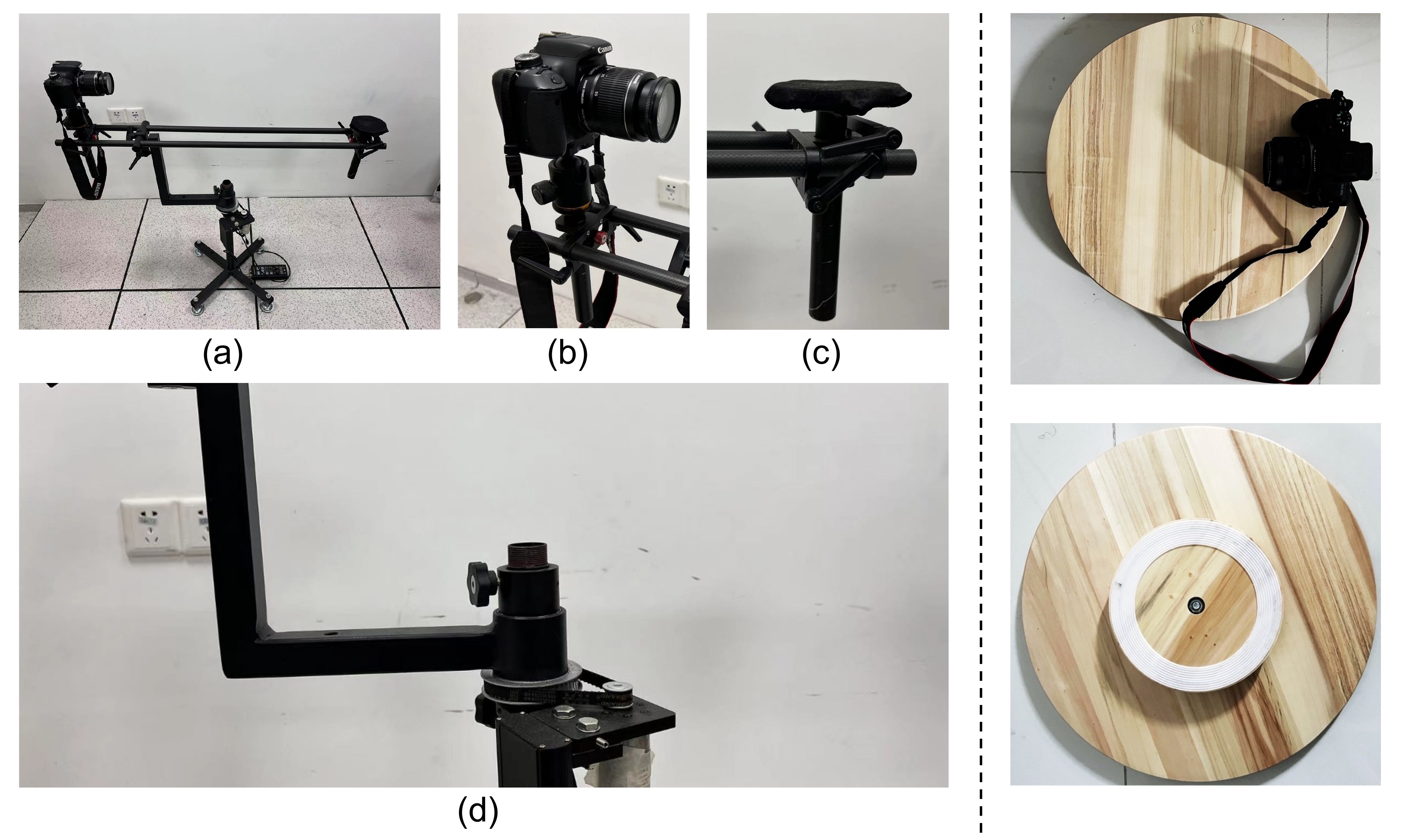"}
    \caption{Left: (a) Overview of the device, (b) Stand for the camera, (c) Stand for the object being photographed with dark cloth for interreflection removal, (d) Rotating hinge. Right: A portable version of image capturing device, shown in top and bottom views.}
    \label{fig: device}
\end{figure}

\noindent\textbf{Device introduction.}  {\sc Soldier}, {\sc Player}, {\sc Policeman}, and {\sc Dancer}'s observed images were captured by a customized device shown in \Fref{fig: device} (left), which consists of two stands (one for holding the subject being photographed, the other for supporting the camera) and a rotating mechanism. The distance from the camera to the object is adjustable. In addition to this, we also consider a more portable device shown in \Fref{fig: device} (right), which is made up of a wooden rotatable platform\footnote{\url{https://www.ikea.com/sg/en/p/snudda-lazy-susan-solid-wood-40176460/}} with a diameter of 39mm and the camera. We capture {\sc Eevee}'s observed images based on this device.

\noindent\textbf{Photographing requirements.} Before photographing, the distance between the camera and the object is determined based on the proportion of the object in the viewfinder, ensuring a balance of the occupied portion between the objects and the camera. Three typical distances were used: 0.9 meters for large and 0.4 meters (or 0.3 meters) for small objects. During photographing, the thumb rule is to capture a clear image with less noise and keep rotation velocity as uniform as possible. For the camera's parameters, we chose ISO 1600, an aperture size of f/13 for outdoor scenes; and ISO 3200, an aperture size of f/6.3 for indoor scenes, respectively. The focal size is fixed at 31mm for different scenes.

\noindent \textbf{Pre-processing pipeline}. In the pre-processing pipeline, we extracted 50 images from the video at equal intervals to use as our data. We then obtain the objects' masks in each scene from the first frame by Photoshop. Those masks help separate objects and backgrounds. In practice, there are translational motions in the horizontal and vertical directions, mostly obvious on objects due to structural instability. Therefore, after calculating the relative rotation angle $\theta_j$, we used a simple algorithm for motion correction, assuming that the only motion of the object relative to the camera was translational in the horizontal and vertical directions. Specifically, we pre-set the range of motion and iterate over the distance vector to find the distance of movement (plus or minus 20 pixels) that minimizes the difference between the front and back frames after applying the mask. Note that although large movement is corrected in this step, minor movements still exist and are hard to eliminate. Fortunately, our method can tolerate those minor movements.

\clearpage
\newpage
\section{Qualitative Comparison}
\label{sec:qualitative comparison}
\subsection{Qualitative Comparison on Synthetic Dataset}
\begin{figure}[t]
    \centering
    \includegraphics[width=0.73\textwidth]{"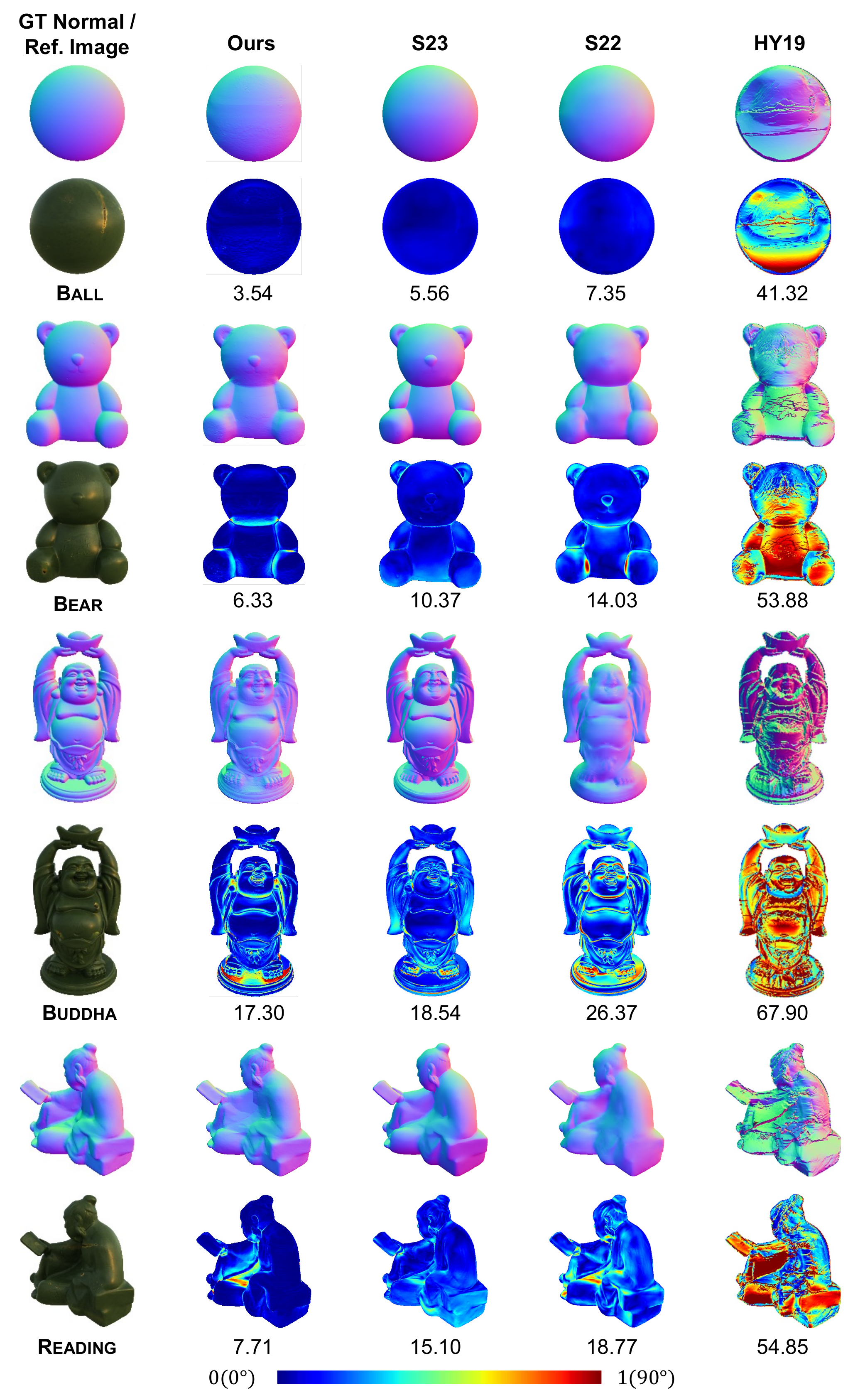"}
    \caption{The visual quality comparison among Spin-UP, S23~\cite{ikehata2023scalable}, S22~\cite{ikehata2022universal}, and HY19~\cite{haefner2019variational} on  the shape group in terms of normal map (row 1, 3, 5, 7), error map (row 2, 4, 6, 8). Numbers indicate the MAE for surface normal.}
    \label{fig:shape}
\end{figure}

\begin{figure*}[t]
    \centering
    \includegraphics[width=0.70\textwidth]{"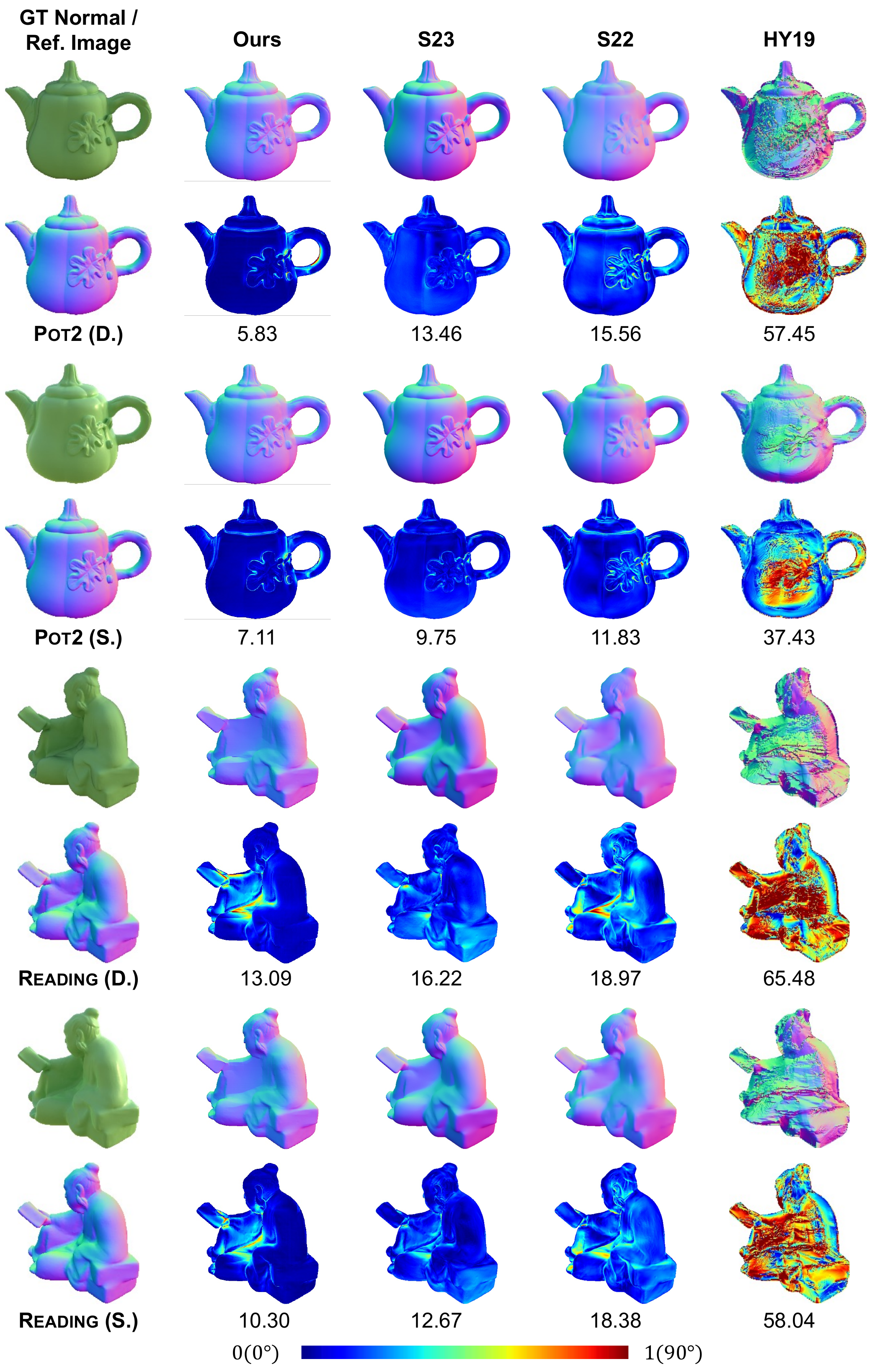"}
    \caption{The visual quality comparison among Spin-UP, S23~\cite{ikehata2023scalable}, S22~\cite{ikehata2022universal}, and HY19~\cite{haefner2019variational} on  the reflectance group in terms of normal map (row 1, 3, 5, 7), error map (row 2, 4, 6, 8). Numbers indicate the MAE for surface normal.}
    \label{fig:reflectance}
\end{figure*}

\begin{figure*}[t]
    \centering
    \includegraphics[width=0.70\textwidth]{"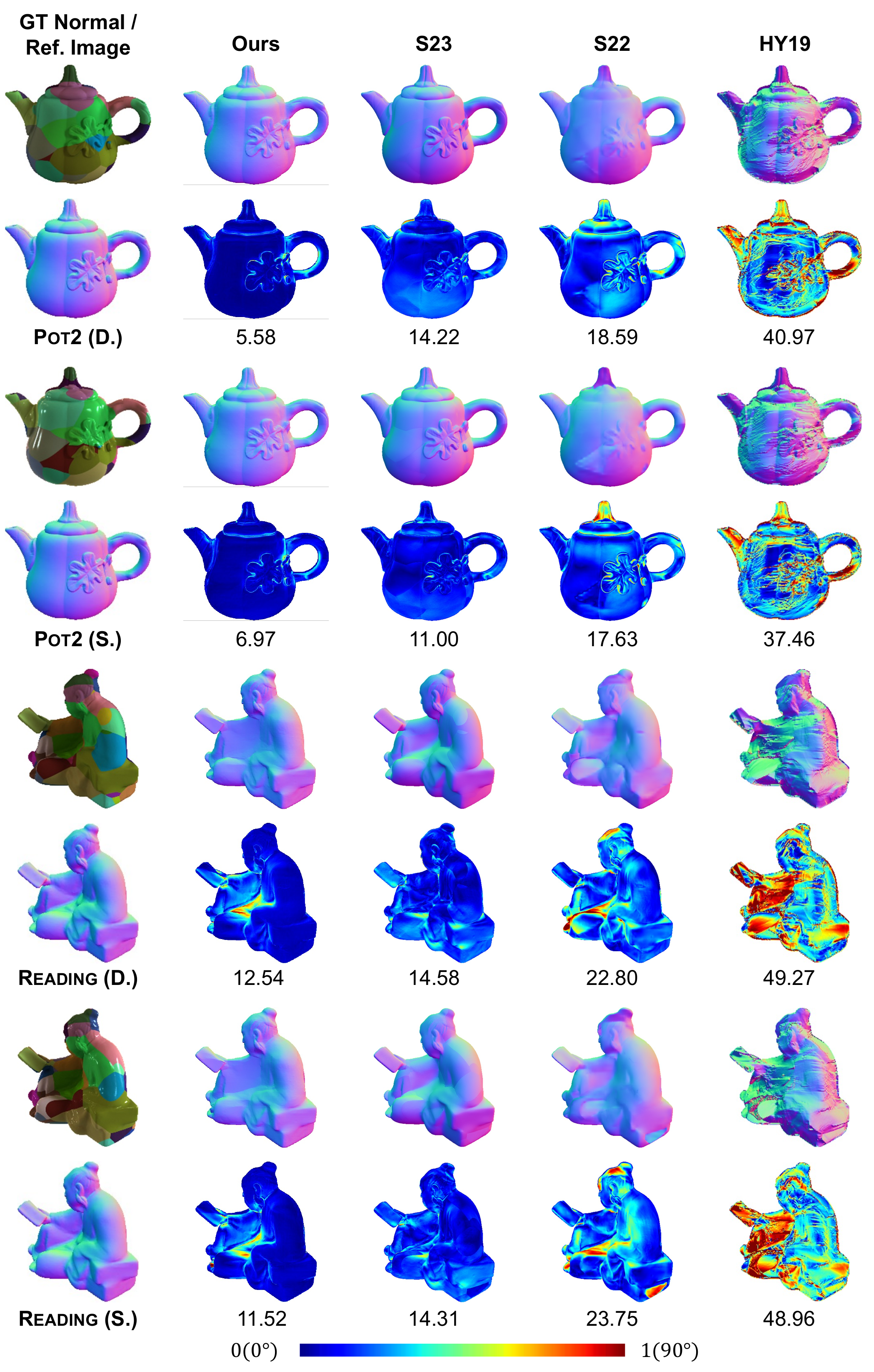"}
    \caption{The visual quality comparison among Spin-UP, S23~\cite{ikehata2023scalable}, S22~\cite{ikehata2022universal}, and HY19~\cite{haefner2019variational} on  the spatially varying material group in terms of normal map (rows 1, 3, 5, 7), error map (rows 2, 4, 6, 8). Numbers indicate the MAE for surface normal.}
    \label{fig:sv-material}
\end{figure*}

We show all the estimated normal maps, error maps of Spin-UP, S23~\cite{ikehata2023scalable}, S22~\cite{ikehata2022universal}, and HY19~\cite{haefner2019variational} of shape, light, reflectance, and spatially-varying material groups in~\Fref{fig:shape}-\Fref{fig:sv-material}.
\begin{figure}[H]
    \centering
    \includegraphics[width=0.71\textwidth]{"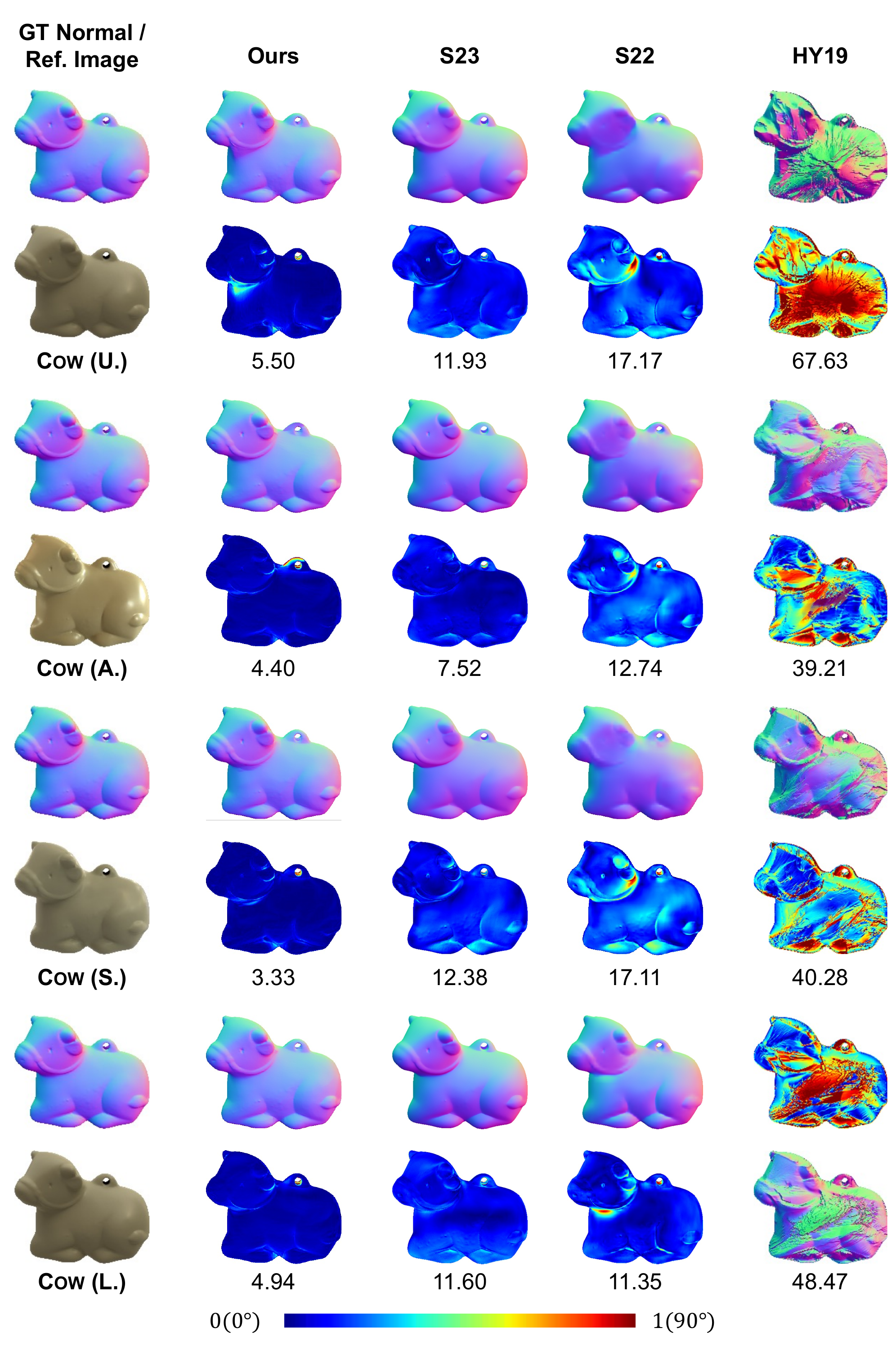"}
    \caption{The visual quality comparison among Spin-UP, S23~\cite{ikehata2023scalable}, S22~\cite{ikehata2022universal}, and HY19~\cite{haefner2019variational} on  the light group in terms of normal map (row 1, 3, 5, 7), error map (row 2, 4, 6, 8). Numbers indicate the MAE for surface normal.}
    \label{fig:light}
\end{figure}

\clearpage
\newpage

\clearpage
\newpage

\clearpage
\newpage

\clearpage
\newpage
\subsection{Qualitative Comparison on Real-world Dataset}
\begin{figure}[H]
    \centering
    \includegraphics[width=0.9\textwidth]{"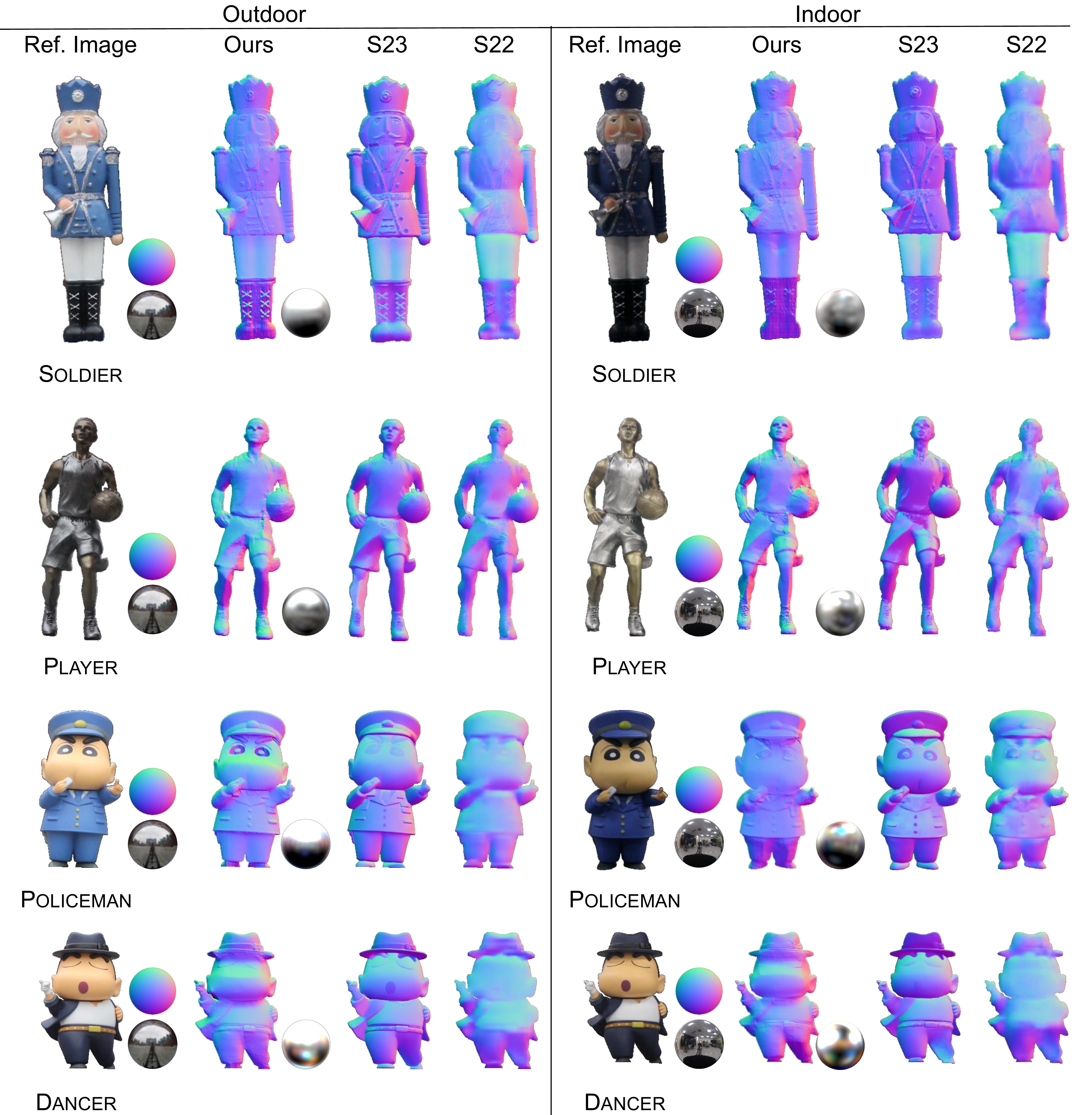"}
    \caption{The visual quality comparison among Spin-UP, S23~\cite{ikehata2023scalable}, and S22~\cite{ikehata2022universal} on the {\sc Soldier, Player, Policeman}, and {\sc Dancer} in terms of the normal map. Left (right) side of the solid line: objects captured in {\sc Campus} ({\sc Workplace}) environment.}
    \label{fig:real_1}
\end{figure}

\begin{figure}[t]
    \centering
    \includegraphics[width=1.\textwidth]{"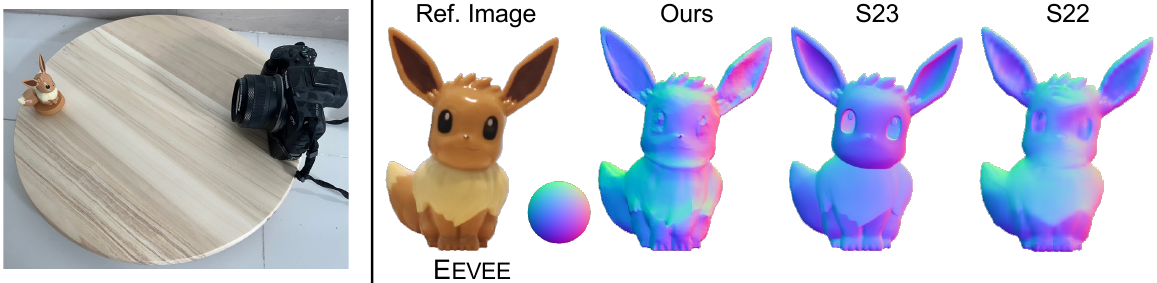"}
    \caption{The visual quality comparison among Spin-UP, S23~\cite{ikehata2023scalable}, and S22~\cite{ikehata2022universal} on {\sc Eevee} captured in a living room in terms of the normal map based on more portable device.}
    \label{fig:real_2}
\end{figure}

We show all the estimated normal maps of Spin-UP, S23~\cite{ikehata2023scalable}, and S22~\cite{ikehata2022universal} of real-world dataset in~\Fref{fig:real_1} and ~\Fref{fig:real_2}.

\newpage
\end{document}